\documentclass[nohyperref]{article}

\usepackage{microtype}
\usepackage{graphicx}
\usepackage{subfigure}
\usepackage{booktabs} % for professional tables

\usepackage{amsmath,amsfonts,amsthm,amssymb,bm}

\def\eqref#1{equation~\ref{#1}}
\def\1{\bm{1}}

\def\rvepsilon{{\boldsymbol{\epsilon}}}
\def\rvtheta{{\boldsymbol{\theta}}}

\def\vb{{\bm{b}}}

\def\vm{{\bm{m}}}

\def\vt{{\bm{t}}}
\def\vu{{\bm{u}}}

\def\vw{{\bm{w}}}
\def\vx{{\bm{x}}}
\def\vy{{\bm{y}}}

\def\mA{{\bm{A}}}

\def\mJ{{\bm{J}}}

\def\mM{{\bm{M}}}

\def\mQ{{\bm{Q}}}

\def\mU{{\bm{U}}}
\def\mV{{\bm{V}}}
\def\mW{{\bm{W}}}

\DeclareMathAlphabet{\mathsfit}{\encodingdefault}{\sfdefault}{m}{sl}
\SetMathAlphabet{\mathsfit}{bold}{\encodingdefault}{\sfdefault}{bx}{n}
\newcommand{\tens}[1]{\bm{\mathsfit{#1}}}

\def\tM{{\tens{M}}}

\usepackage{tikz}
\usetikzlibrary{positioning}
\usetikzlibrary{fit}
\usetikzlibrary{calc}

\usepackage{enumitem}
\setlist{leftmargin=2.5mm}

\usepackage{hyperref}

\usepackage[accepted]{icml2022}

\usepackage{amsmath}
\usepackage{amssymb}
\usepackage{mathtools}
\usepackage{amsthm}

\usepackage[capitalize]{cleveref}

\theoremstyle{plain}
\newtheorem{theorem}{Theorem}[section]
\newtheorem{proposition}[theorem]{Proposition}

\theoremstyle{definition}
\newtheorem{definition}[theorem]{Definition}

\theoremstyle{remark}

\usepackage[textsize=tiny]{todonotes}

\icmltitlerunning{A Data-Augmentation Is Worth A Thousand Samples}

\begin{document}

\twocolumn[
\icmltitle{A Data-Augmentation Is Worth A Thousand Samples:\\Exact Quantification From Analytical Augmented Sample Moments}




\begin{icmlauthorlist}
\icmlauthor{Randall Balestriero}{yyy}
\icmlauthor{Ishan Misra}{yyy}
\icmlauthor{Yann LeCun}{yyy}
\end{icmlauthorlist}

\icmlaffiliation{yyy}{Meta/Facebook AI Research}

\icmlcorrespondingauthor{Randall Balestriero}{rbalestriero@fb.com}

\icmlkeywords{Machine Learning, ICML}

\vskip 0.3in
]



\printAffiliationsAndNotice{\icmlEqualContribution} % otherwise use the standard text.

\begin{abstract}
Data-Augmentation (DA) is known to improve performance across tasks and datasets. We propose a method to theoretically analyze the effect of DA and study questions such as: how many augmented samples are needed to correctly estimate the information encoded by that DA? How does the augmentation policy impact the final parameters of a model? We derive several quantities in close-form, such as the expectation and variance of an image, loss, and model's output under a given DA distribution. Those derivations open new avenues to quantify the benefits and limitations of DA. For example, we show that common DAs require tens of thousands of samples for the loss at hand to be correctly estimated and for the model training to converge. We show that for a training loss to be stable under DA sampling, the model's saliency map (gradient of the loss with respect to the model's input) must align with the smallest eigenvector of the sample variance under the considered DA augmentation, hinting at a possible explanation on why models tend to shift their focus from edges to textures.
\end{abstract}

\section{Introduction}

Data augmentation is a prevalent technique in training deep learning models.
These models $f_{\theta}$, governed by some parameters $\theta \in \Theta$, are trained on the train set and expected to generalize to unseen samples (test set).
Data Augmentation (DA) serves to improve this generalization behavior of the models.
Assuming that the space $\mathcal{F}\triangleq \{f_{\theta}:\forall \theta \in \Theta\}$ is diverse enough, theoretically, an accurate DA policy is all that is needed to close any performance gap between the train and test sets \cite{unser2019representer,zhang2021understanding}.

\begin{figure}[H]
    \centering
\begin{tikzpicture}[%
    auto,
    block/.style={
      rectangle,
      draw=blue,
      thick,
      fill=blue!20,
      text width=5em,
      align=center,
      rounded corners,
      minimum height=2em
    }]

\node[inner sep=0pt] at (0,0) (sampled)
    {\includegraphics[width=0.4\linewidth]{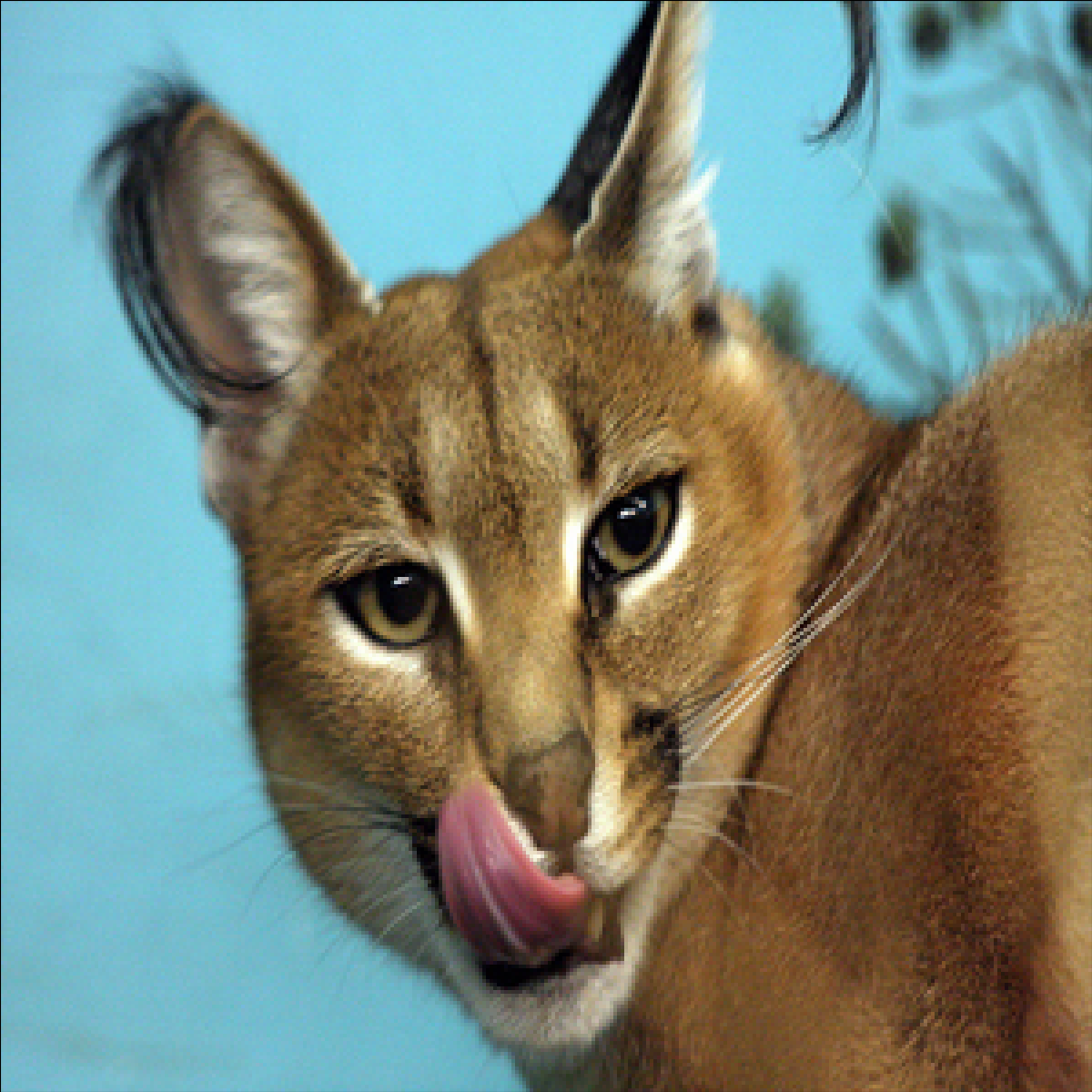}};
\node[align=center, above=0.0cm of sampled,inner sep=0.01cm] {\parbox{2.2cm}{\centering\small{\bf original image $I$}}};

\node[draw=blue,line width=1mm,inner sep=0pt] (matrix) at (-2,-3.8)
    {\includegraphics[width=0.4\linewidth]{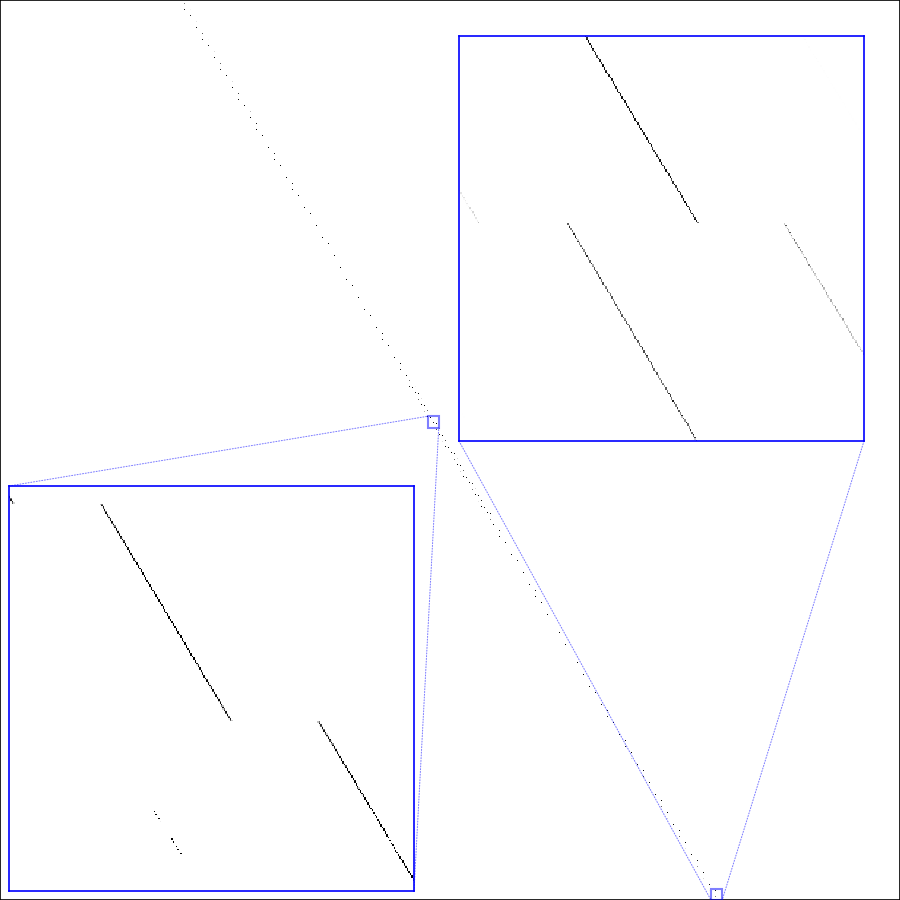}};
\node[draw=red,line width=1mm,inner sep=0pt, right=0.1cm of matrix] (vector)
    {\includegraphics[width=0.02\linewidth,height=3.3cm]{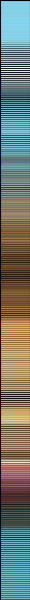}};
\node[align=center, above=0.03cm of matrix,inner sep=0cm] {\small\parbox{3.5cm}{\centering {\bf zoom $\theta=0.6$}}};  

\node[draw=blue,line width=1mm,inner sep=0pt] (matrix) at (2,-3.8)
    {\includegraphics[width=0.4\linewidth]{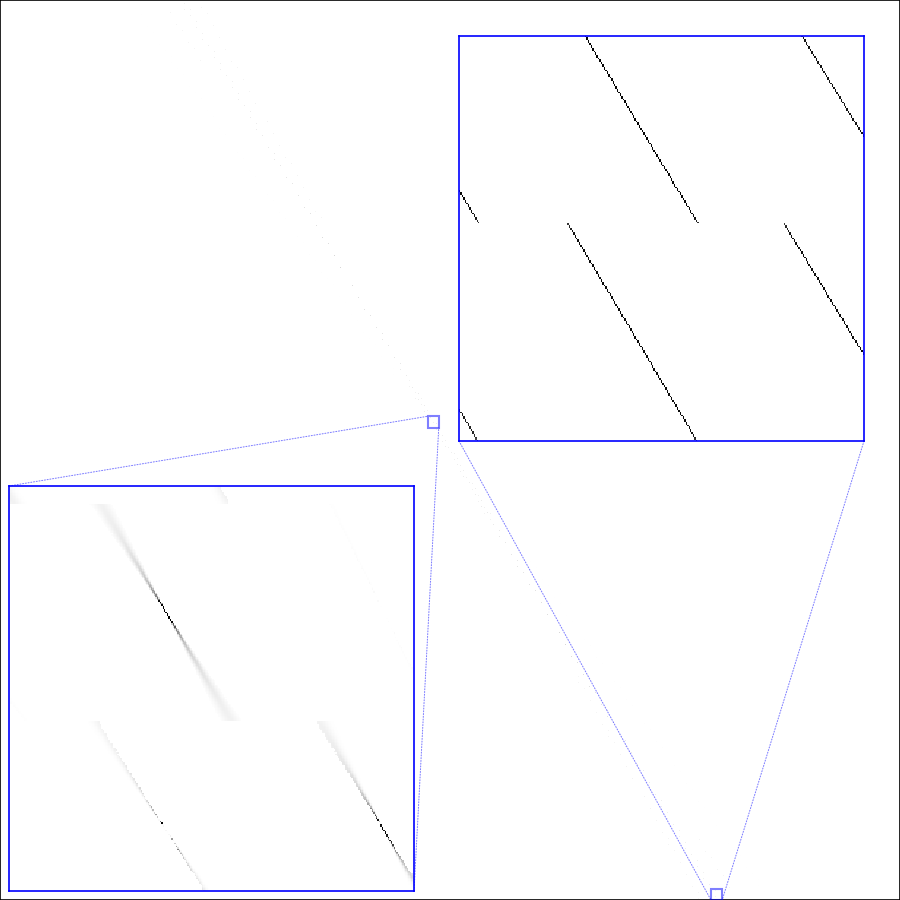}};
\node[draw=red,inner sep=0pt,line width=1mm, right=0.1cm of matrix] (vector)
    {\includegraphics[width=0.02\linewidth,height=3.3cm]{mainfigure/vector_image.png}};
\node[align=center, above=0.0cm of matrix,inner sep=0cm] {\small\parbox{3.5cm}{\small\centering {\bf zoom $\theta\sim \mathcal{N}(0.6,0.05)$}}};

\node[inner sep=0pt] (transformed) at (-2,-7.5)
    {\includegraphics[width=0.4\linewidth]{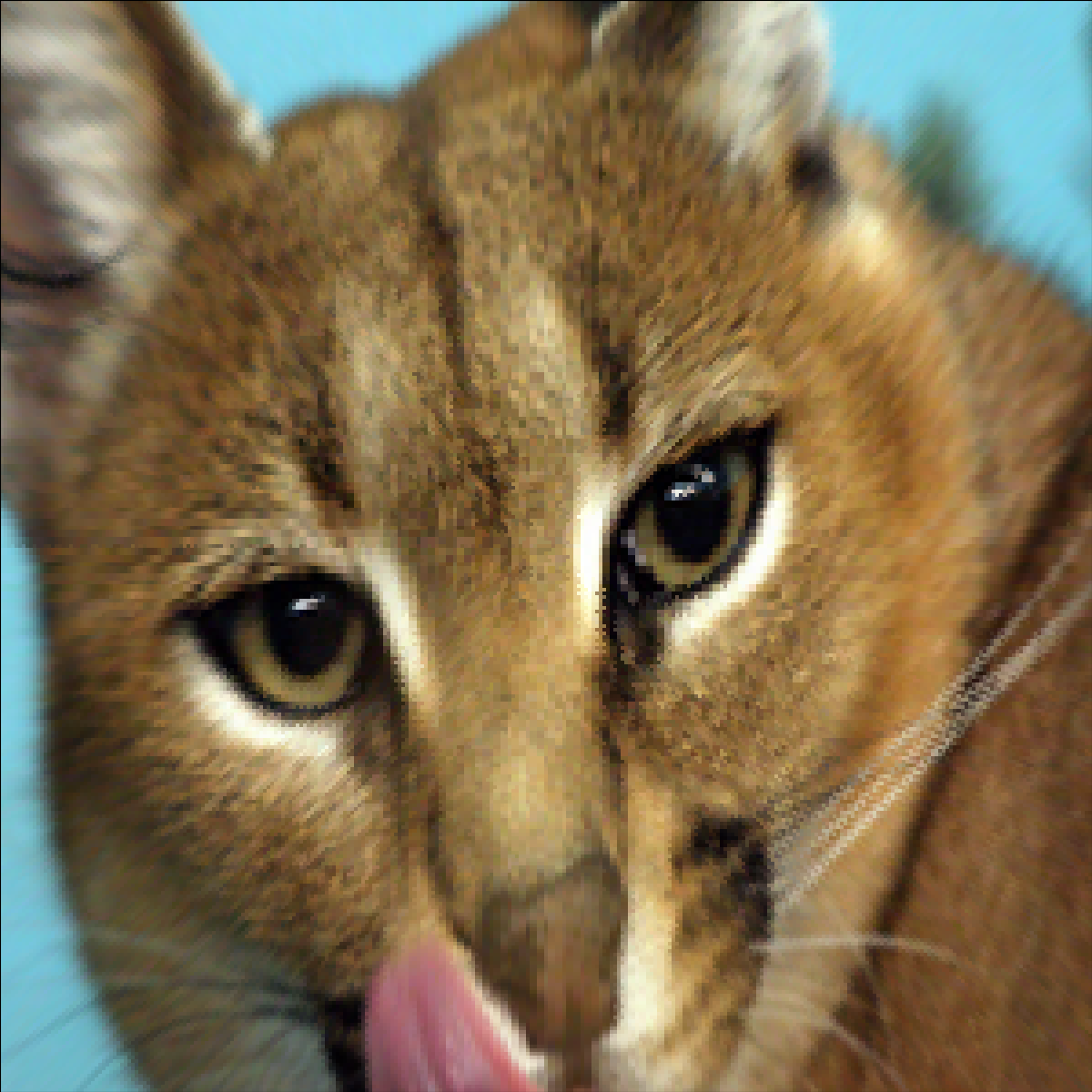}};
\node[align=center, below=0.05cm of transformed] {\small\parbox{3.5cm}{\centering {\bf transformed image $I(\theta)$}}};

\node[inner sep=0pt] (transformedsmooth) at (2,-7.5)
    {\includegraphics[width=0.4\linewidth]{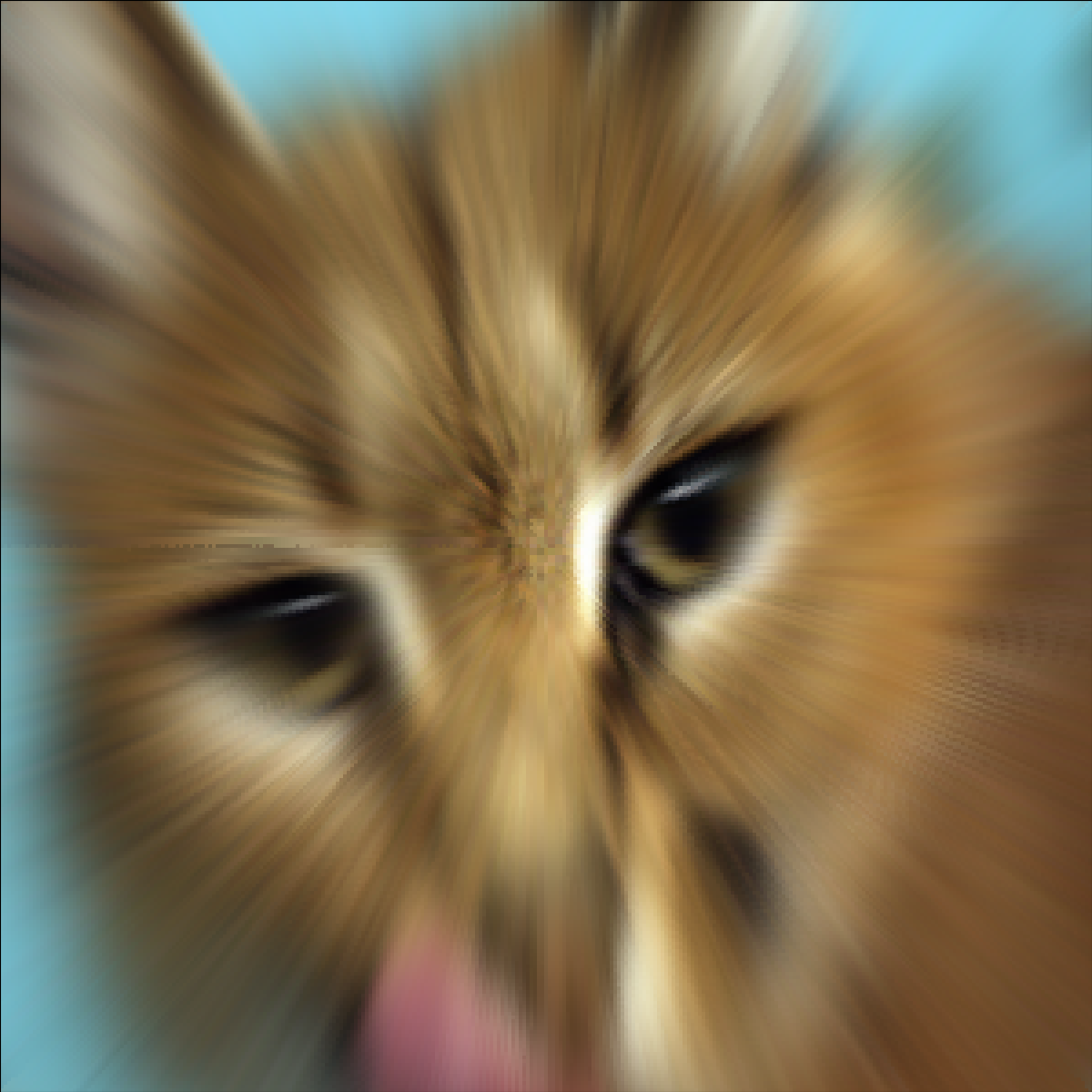}};
\node[align=center, below=0.05cm of transformedsmooth] {\small\parbox{3.5cm}{\centering {\bf expected image $\mathbb{E}_{\theta}[I(\theta)]$}}};

% \node[inner sep=0pt,above=0.8cm of matrix] (closest)
%     {\includegraphics[width=0.2\linewidth]{mainfigure/closest_image.png}};

% \node[align=left, right=0.03cm of vector,minimum width=5cm] (text1) {\small
% {\bf
% Data-space transform (ours). (\cref{eq:basis_image_model})}\\
% \parbox{5cm}{\small the image $I$ is flattened, projected through a matrix $\mM$ (possibly per color channels), and reshaped to produce $T$. The rows of $\mM$ are the flattened generalized function $h(i,j,.,.)$ whose role is to extract the needed value(s) of $I$ for each desired coordinate of $T$.
% }
% };

% \node[align=left, right=0.3cm of closest,minimum width=5cm] (text2) {\small
%     {\bf Coordinate-space transform. (\cref{eq:coordinate_transform})}\\
%     \parbox{5cm}{\small each coordinate $(i,j)$ of $T$  corresponds to the coordinate $t(i,j)$ of $I$
% ({\bf black dots}). Since those are not observed, some flavor of interpolation
% is employed to estimate them based on the observed coordinates of $I$ ({\bf white dots}).}
% };

% \node[draw,fit=(matrix.south west) (matrix.north west) (text1.north east),black!70,rounded corners=0.2cm,very thick,fill=black!10,fill opacity=0.2] (box1) {};

% \node[draw,fit=(closest.south west) (closest.north west) (text2.north east),black!70,rounded corners=0.2cm,very thick,fill=black!10,fill opacity=0.2] (box2) {};

\draw [dotted,very thick](0.16,-2.1)--(0.16,-9.1);
\draw [->,very thick](0,-1.7)--(-0.3,-1.95);
\draw [->,very thick](0,-1.7)--(0.3,-1.95);
\draw [->,very thick](-2,-5.6)--(-2,-5.8);
\draw [->,very thick](2,-5.6)->(2,-5.8);
% \node (or) [fill=white] at ($(box1)!0.5!(box2)$) {};

% \draw [->,very thick] (sampled.west) to (box1.east);
% \draw [->,very thick] (sampled.west) to (box2.east); 
% \draw [->,very thick] (box1.west) to (transformed.east);
% \draw [->,very thick] (box2.west) to (transformed.east) ; 

\end{tikzpicture}
\vspace{-0.5cm}
    \caption{
    We propose a novel way to express Data Augmentation (DA) analytically that allows us to understand the impact of DA on the learned parameters of a model and quantify DA's sample efficiency.
    We model the transformation of an image through a matrix-vector multiplication of the flattened image ({\color{red}red}) and a sparse matrix ({\color{blue}blue}) representing the transform.
    This allows us to compute the analytical expectation and variance of the transformed data (or any function computed with it) in closed form with respect to the transformation parameters.}
    % We define a novel pixel-space transformation model (\cref{sec:new_transformation}) that performs any desired transformation of an image ({\bf first row}) through matrix-vector multiplication between the \color{red}flattened image\color{black}, and a \color{blue}{\em sparse} matrix \color{black} ({\bf second row}) to produce the transformed image ({\bf bottom left}).
    % By carefully designing that matrix (\cref{def:our_transform,thm:affine_explicit}), we allow the {\bf analytical expectation and variance of the transformed data (or a loss/model output being a function of it) to be obtained in closed-form when taken with respect to the considered transformation parameters} ({\bf bottom right, more examples in  \cref{fig:expectation}}).
    % The analytical form of the expected loss and transformed image open new avenues such as understanding the impact of DA into the learned parameters (\cref{sec:motivation}) or quantifying the sample efficiency of DA sampling (\cref{sec:monte_carlo}).
    % }
\vspace{-0.5cm}
    \label{fig:summary}
    % \vspace{-0.8cm}
\end{figure}

While there are multiple ways to maximize test set performance in a finite data regime, such as restricting the space of models $\mathcal{F}$, adding regularization on $f_{\theta}$ etc., DA continues to play a critical role for good performance \citep{perez2017effectiveness,taqi2018impact,shorten2019survey}.
Some regimes such as self-supervised learning (SSL), rely entirely on augmentations \citep{misra2020self,zbontar2021barlow}.
Despite its empirical effectiveness, our understanding of DA has many open questions, three of which we propose to study: \textbf{(a)} how do different DAs impact the model's parameters during training?; \textbf{(b)} how sample-efficient is the DA sampling, i.e., how many DA samples a model must observe to converge?; and \textbf{(c)} how sensitive is a loss/model to the DA sampling and how this variance evolves during training as a function of the model's ability to minimize the loss at hand, and as a function of the model's parameters? 

Our goal is to analytically derive some preliminary answers and insights into those three areas enabled by a novel image transformation operator that we introduce in \cref{sec:new_transformation} coined Data-Space Transform (DST).
We summarize our \textbf{contributions} below:
\begin{enumerate}[leftmargin=0.55cm,label=(\alph*),topsep=-4pt,itemsep=-1ex]
\item we derive the analytical first order moments of augmented samples, and of the losses employing augmented samples (\cref{sec:explicit}), effectively providing us with the explicit regularizer induced by each DA (\cref{sec:motivation})
\item we quantify the number of DA samples that are required for a model/loss to obtain a correct estimate of the information conveyed by that DA (\cref{sec:monte_carlo})
\item we derive the sensitivity, i.e. variance, of a given loss and model under a DA policy (\cref{sec:sensitivity}) leading us to rediscover from first principles a popular deep network regularizer: TangentProp, as being the natural regularization to employ to minimize the loss variance (\cref{sec:variance}).
\end{enumerate}

\textbf{Upshot of gained insights:} \textbf{(a)} shows us that the explicit DA regularizer corresponds to a generalized Tikhonov regularizer that depends on each sample's covariance matrix largest eigenvectors; the kernel space of the model's Jacobian matrix aligns with the data manifold tangent space as modeled by the DA \textbf{(b)} shows us that the number of augmented samples required for a loss/model to correctly estimate the information provided by a DA of a single sample is on the order of $10^4$. Even when considering thousands of samples simultaneously, where each sample's augmentations can complement another sample's augmentations in estimating the information provided by a DA, we find that the entire train set must be augmented at least $50\times$ for a DA policy to be correctly learned by a model \textbf{(c)} 
quantifies the loss variance as a simple function of the model's Jacobian matrix, and the eigenvectors of the augmented sample variance matrix. That is, and echoing our observation from (a), regardless of the model or task-at-hand, the loss sensitivity to random DA sampling goes down as the kernel of a model's Jacobian matrix aligns with the principal directions of the data manifold tangent space; additionally, if one were to concentrate on minimizing the loss sensitivity to DA sampling, TangentProp would be the natural regularizer to employ.

All the proofs and implementation details are provided in the appendix.

\section{Background}
\label{sec:background}

{\bf Explicit Regularizers From Data-Augmentation.}~
It is widely accepted that data-augmentation (DA) regularizes a model towards the transformations that are modeled \citep{neyshabur2014search,neyshabur2017implicit}, and this impacts performances significantly, possibly as much as the regularization offered by the choice of DN architecture \citep{gunasekar2018implicit} and optimizer \citep{soudry2018implicit}.

To gain precious insights into the impact of DA onto the learned functional $f_{\gamma}$, the most common strategy is to derive the {\em explicit} regularizer that directly acts upon $f_{\gamma}$ in the same manner as if one were to use DA during training.
This explicit derivation is however challenging and so far has been limited to DA strategies such as additive white noise or multiplicative binary noise applied identically and independently throughout the image, as with dropout \citep{srivastava2014dropout,bouthillier2015dropout}.
In those settings, various works have studied in the linear regime the relation between such data-augmentation and its equivalence to using Tikhonov \citep{tikhonov1943stability} or weight decay \citep{zhang2018three} as in $\min_{\mW} \sum_{n=1}^{N} \mathbb{E}_{\rvepsilon \sim \mathcal{N}(0,\sigma)}\left[\| \vy_n-\mW(\vx+\rvepsilon)\|_2^2\right] =\min_{\mW} \sum_{n=1}^{N} \| \vy_n-\mW\vx\|_2^2 + \lambda(\sigma) \|\mW\|_F^2$ \citep{baldi2013understanding}. More recently, \citet{wang2018max,lejeune2019implicit} extended the case of additive white noise to nonlinear models and concluded that this DA corresponds to adding an explicit Frobenius norm regularization onto the Jacobian of $f_{\theta}$ evaluated at each data sample.

Going to more involved DA strategies e.g. translations or zooms of the input images is challenging and has so far only been studied from an empirical perspective. For example, \citet{hernandez2018deep,hernandez2018data,hernandez2018further} performed thorough ablation studies on the interplay between DAs and a collection of known explicit regularizers to find correlations between them. It was concluded that weight-decay (the explicit regularizer of additive white noise) does not relate to those more advanced DAs. This also led other studies to suggest that norm-based regularization might be insufficient to describe the implicit regularization of DAs involving advanced image transformations \citep{razin2020implicit}. We debunk this last claim in  Sec.~\ref{sec:explicit}.

{\bf Coordinate Space Transformation.}~
Throughout this paper, we will consider a two-dimensional image $I(x,y)$ to be at least square-integrable $I \in L^2(\mathbb{R}^2)$ \citep{mallat1989multifrequency}. Multi-channel images are dealt with by applying the same transformation on each channel, as commonly done in practice \citep{goodfellow2016deep}. As we are interested in practical cases, we will often assume that $I$ has compact support e.g. has nonzero values only within a bounded domain such as $[0,1]^2$. Visualizing this image thus corresponds to displaying the sampled values of $I$ on a regular grid (pixel positions) of $[0,1]^2$ \citep{heckbert1982color}.

The most common formulation to apply a transformation on the image $I$ to obtain the transformed image $T$ is to transform the image coordinates \citep{sawchuk1974space,wolberg2000robust,mukundan2001image}. That is, a mapping $t:\mathbb{R}^2 \mapsto \mathbb{R}^2$ describes what coordinate $t(u,v)$ of $I$ maps to the coordinate $u,v$ of $T$ as in 
\begin{equation}
  T(u,v)=I(t(u,v))  .\label{eq:coordinate_transform}
\end{equation}
This function $t$ often has some parameter $\theta$ governing the underlying transformation as in
\begin{align*}
    t_{\theta}(x,y)=&[x-\theta_1,y-\theta_2]^T,&&\text{(translation)}\\
    t_{\theta}(x,y)=&\begin{bmatrix}
    \cos(\theta) & -\sin(\theta)\\
    \sin(\theta) & \cos(\theta)
    \end{bmatrix}\begin{bmatrix}
    x\\
    y
    \end{bmatrix},&&\text{(rotation)}\\
    t_{\theta}(x,y)=&[\theta_1x,\theta_2y]^T,&&\text{(zoom)}
\end{align*}
We provide a visual depiction of the zoom transformation applied in coordinate space in Fig.~\ref{fig:coordinate}, in the appendix. The formulation of Eq.~\ref{eq:coordinate_transform} has two key benefits. First, it allows a simple and intuitive design of $t$ to obtain novel transformations. Second, it is computationally efficient as the coordinate-space of images are $2/3$-dimensional. Those benefits have led to the design of deep network architecture with explicit coordinate transformations being embedded into their layers, as with the Spatial Transformer Network \citep{jaderberg2015spatial}. On the other hand, Eq.~\ref{eq:coordinate_transform} has one major drawback for our purpose: computing the exact moments of the transformed image under random $\theta$ parameters (e.g. the expectation $\mathbb{E}_{\rvtheta}[I\circ t_{\rvtheta}]$) is not tractable due to the composition of $t$ with the nonlinear mapping $I$. And as it will become clear in \cref{sec:monte_carlo_motivation}, computing such quantities is crucial to better grasp the many properties around the use of DA sampling during training e.g. to study its impact onto a model's parameters.

\section{Analytical Moments of Transformed Images Enable Infinite Dataset Training}
\label{sec:rethinking}

We first motivate this study by formulating the training process under DA sampling as doing a Monte-Carlo estimate of the true (unknown) expected loss under that DA distribution (\cref{sec:monte_carlo_motivation}). Going beyond this sampling/estimation procedures requires knowledge of the expectation and variance of a transformed sample, motivating the construction of a novel Data-Space Transformation (DTS) (\cref{sec:new_transformation}) that allows for a closed-form formula of those moments (\cref{sec:explicit}). From those results, we will be able to remove the need to sample transformed images to train a model, by obtaining the closed-form expected loss \cref{sec:motivation}.

\begin{figure*}[h!]
    \centering
    \includegraphics[width=\linewidth]{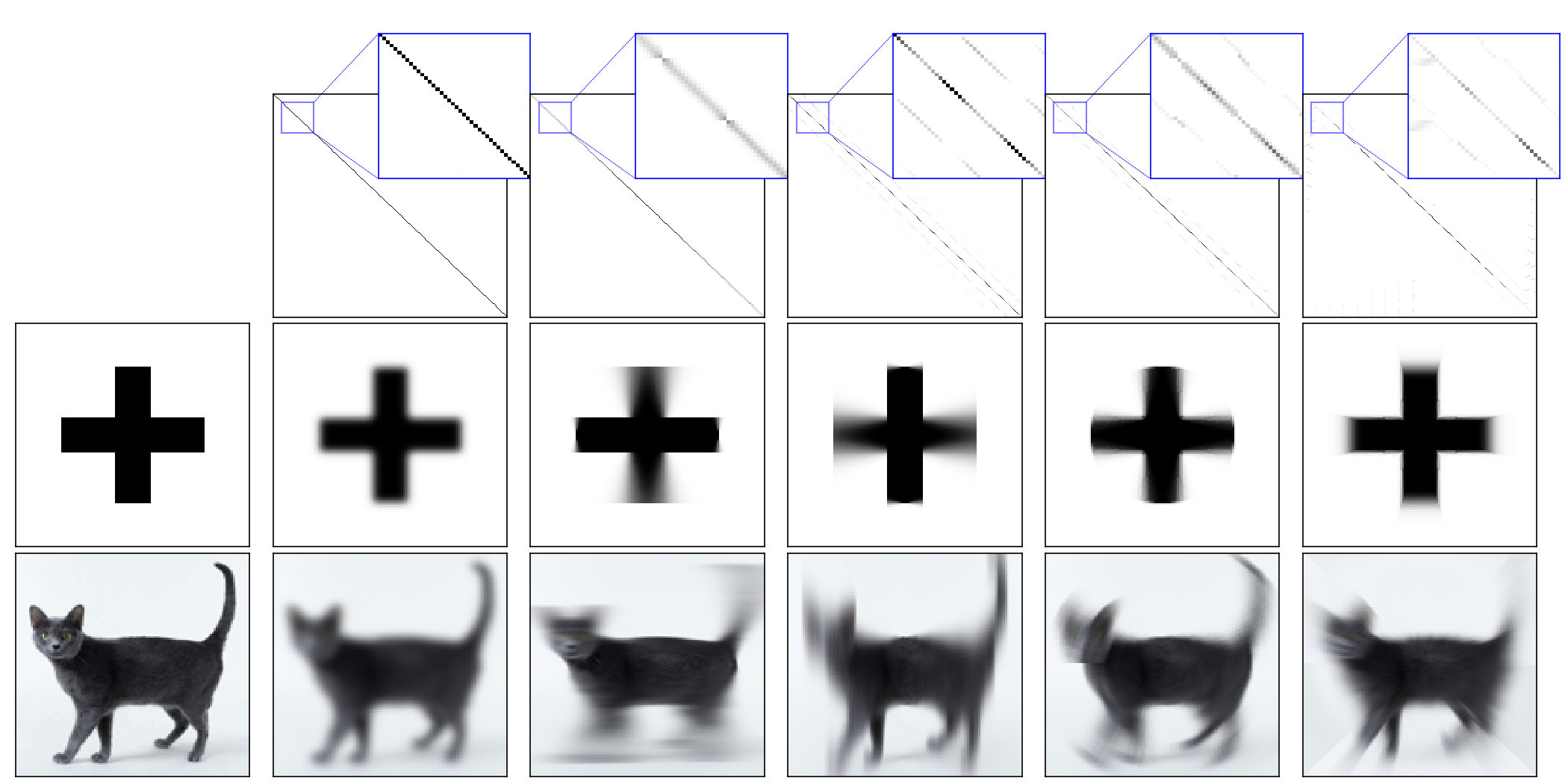}\\
    \begin{minipage}{0.16\linewidth}\centering
    {\small original}
    \end{minipage}
    \begin{minipage}{0.16\linewidth}\centering
    {\small horizontal+vertical}\\
    {\small translation}
    \end{minipage}
    \begin{minipage}{0.16\linewidth}\centering
    {\small horizontal shear}
    \end{minipage}
    \begin{minipage}{0.16\linewidth}\centering
    {\small vertical shear}
    \end{minipage}
    \begin{minipage}{0.16\linewidth}
    \centering
    {\small rotation}
    \end{minipage}
    \begin{minipage}{0.16\linewidth}
    \centering
    {\small zoom}
    \end{minipage}
    \vspace{-0.3cm}
    \caption{{\bf Top row:} matrix operator $\mathbb{E}_{\theta}[\mM(\theta)]$ producing the expected image via $\mathbb{E}_{\theta}[\vt(\theta)]=\mathbb{E}_{\theta}[\mM(\theta)] \vx$ (recall \cref{eq:discrete} and the discussion below \cref{thm:affine_explicit}), bottom block in \cref{fig:summary}) that transforms a (flattened) image into the average (expected) image after taking the expectation with respect to a given transformation distribution (as in \cref{eq:expectation_example}), here with distributions: $\mathcal{N}(0,0.04)\otimes \mathcal{N}(0,0.04)$ for translation, $\mathcal{N}(0,0.2)$ for both shearing, $\mathcal{N}(0,0.1)$ for rotation and $\mathcal{N}(1,0.1)$ for zoom. This \textit{ sparse} matrix is the same regardless of the image and is derived in closed-form for the above standard transformations (\cref{thm:affine_explicit}). {\bf Middle \& Bottom row:} are two sample images (original on the left) that  depict for each column the ``expected image'' under each transformation. Estimating those images (and any loss employing them) only through sampling the DA augmentations would require tens of thousands of samples (see \cref{fig:convergence_translation}).}
    \label{fig:expectation}
\end{figure*}

\subsection{Motivation: Current Data-Augmentation Training Performs Monte-Carlo Estimation}
\label{sec:monte_carlo_motivation}

Training a model with DA consists in (i) sampling transformed images at each training iteration for each sample $\vx_n$ as in $\mathcal{T}_{\theta_n}(\vx_n)$ with $\theta_n\sim \rvtheta$ a randomly sampled DA parameter e.g. the amount of translation to apply, (ii) evaluating the loss $\mathcal{L}$ on the transformed sample/mini-batch/dataset, and (iii) using some flavor of gradient descent to update the parameters $\gamma$ of the model $f_{\gamma}$. 

This training procedure corresponds to a one-sample Monte-Carlo (MC) estimate \citep{metropolis1953equation,hastings1970monte} of the expected loss 
\begin{align}
    \sum_{n=1}^{N}\mathbb{E}_{\rvtheta}\left[\mathcal{L}\left(f_{\gamma}(\mathcal{T}_{\rvtheta}(\vx_n)\right)\right]\approx \sum_{n=1}^{N}\mathcal{L}(f_{\gamma}(\mathcal{T}_{\theta_n}(\vx_n)),\label{eq:original}
\end{align}
with $\theta_n$ i.i.d samples from $\rvtheta$. In a supervised setting, the loss would also receive a per-sample target $\vy_n$ as input. Although a one-sample estimate might be insufficient to apply the central limit theorem \citep{rosenblatt1956central} and guarantee training convergence, the combination of multiple samples in each mini-batch and the repeated i.i.d sampling at each training step does provide convergence in most cases. For example, Self-Supervised losses i.e. losses that heavily rely on DA, tend to diverge is the mini-batch size is not large enough, as opposed to supervised losses. To avoid such instabilities, and to incorporate all the DA transformations of $\vx_n$ into the model's parameter update, one would be tempted to compute the model parameters' gradient on the expected loss (left-hand side of \cref{eq:original}) (see \cref{sec:motivation}). Knowledge of the expected loss would also prove useful to measure the quality of the MC estimate (see \cref{sec:monte_carlo}). 

As the closed-form expected loss requires knowledge of the expectation and variance of the transformed sample $\mathcal{T}_{\rvtheta}(\vx_n)$, we first propose to formulate a novel and tractable augmentation model (\cref{sec:new_transformation}) that will allow us to obtain those moments analytically (\cref{sec:explicit}).

\subsection{Proposed Data-Space Transformation}
\label{sec:new_transformation}

Instead of altering the coordinate positions of an image, as done in the coordinate-space transformation of \cref{eq:coordinate_transform}, we propose to alter the image basis functions.

Going back to the construction of functions, one easily recalls that any (image) function can be expanded into its basis as in 
\begin{equation*}
    I(u,v)=\int I(x,y) \delta(u-x,v-y)dx dy,
\end{equation*}
with $\delta$ the usual Dirac distributions. Suppose for now that we consider a horizontal translation by a constant $\theta$. Then, one can obtain that translated image $T$ via
\begin{equation}
    T(u,v)=\int I(x,y) \delta(u-x-\theta,v-y)dx dy,\label{eq:basis_motivation}
\end{equation}
hence by moving the basis functions onto which the image is evaluated, rather than by moving the image's coordinate. As the image is now constant with respect to the transformation parameter $\theta$, and as the basis functions have some specific analytical forms, the derivation of the transformed images expectation and variance, under random $\theta$, will become straightforward. Because the transformed image $T$ is now obtained by combining its pixel values (recall \cref{eq:basis_motivation}) we coin this transform as Data-Space Transform (DST) and we formally define it below.

\begin{definition}[Data-Space Transform]
\label{def:our_transform}
We define the data-space transformation of an image $I\in L^2(\mathbb{R}^2)$ producing the transformed image $T\in L^2(\mathbb{R}^2)$ as
\begin{align}
    T(u,v)=\int I(x,y) h_{\theta}(u,v,x,y)dx dy,\label{eq:basis_image_model}
\end{align}
with $h_{\theta}(u,v,.,.) \in \mathbb{C}^\infty_0(\mathbb{R}^2)$ encoding the transformation.
\end{definition}

In \cref{def:our_transform}, we only impose for $h_{\theta}(u,v,.,.)$ to be with compact support. In fact, one should interpret $h_{\theta}(u,v,.,.)$ as a distribution whose purpose is to evaluate $I$ at a desired (coordinate) position on its domain. This evaluation -depending on the form of $h_{\theta}(u,v,.,.)$- can return a single pixel-value of the image $I$ at a desired location (as in \cref{eq:basis_motivation}), or it can combine multiple values e.g. with $h_{\theta}(u,v,.,.)$ being a bump function. 

{\bf Coordinate-space transformations as DSTs.}
The coordinate-space transformation and the proposed DST (\cref{def:our_transform}) act in different spaces: the image input space and the image output space, respectively. Nevertheless, this does not limit the range of transformations that can be applied to an image. The following statement provides a simple recipe to turn any already employed coordinate-space transformation into a data-space one.

\begin{proposition}
\label{thm:equi}
Any coordinate-space transformation (\ref{eq:coordinate_transform}) using $t:\mathbb{R}^2\mapsto \mathbb{R}^2$ can be expressed as a data-space transformation (\ref{eq:basis_image_model}) by setting $h(u,v,x,y)=\delta(t(x,y)-[u,v]^T)$.
\end{proposition}

Although we will focus here on the standard zoom/rotation/translation/\dots transformations, we propose in \cref{sec:cutmix} a discussion on extending our results to DAs such as CutMix \citep{yun2019cutmix}, CutOut \citep{devries2017improved} or MixUp \citep{zhang2017mixup}. Using \cref{thm:equi}, we obtain the DST operators $h_{\theta}(u,v,x,y)$ to be
\begin{align}
    &\delta(u-x+\theta_1,v-y+\theta_2),\nonumber\\
    &\delta(u-x-\theta_1 y,v-y-\theta_2 x),\nonumber\\
    &\delta(u-\theta x,v-\theta y),\nonumber\\
    &\delta(u-\cos(\theta)x+\sin(\theta)y,v-\sin(\theta)x-\cos(\theta)y),
\label{cor:affine}
\end{align}
for the vertical/horizontal translation, vertical/horizontal shearing, zoom and rotation respectively.
Before focusing on the analytical moments of the data-space transformed samples, we describe how those operators are applied in a discrete setting.

{\bf Discretized version.} We now describe how any DST of a discrete image, flattened as a vector, can be expressed as a matrix-vector product with the matrix entries depending on the employed DA and its parameter $\theta$, as in \cref{fig:summary}. The functional form of the proposed transform in \cref{eq:basis_image_model} producing the target image at a specific position $T(u,v)$ is linear in the image $I$. The continuous integral over the image domain is replaced with a summation with indices based on the desired sampling/resolution of $I$. Expressing linear operators as matrix-vector products will greatly ease our development, we will denote by $\vx \in \mathbb{R}^{hw}$ the flattened $(h\times w)$ discrete images $I$. Hence, our data-space transformation, given some parameters $\theta$, takes the form of 
\begin{equation}
    \vt(\theta) = \mM(\theta) \vx,\label{eq:discrete}
\end{equation}
with $\vt(\theta)\in\mathbb{R}^{hw}$ the flattened transformed image (which can be reshaped as desired) and $\mM(\theta) \in \mathbb{R}^{hw \times hw}$ the matrix whose rows encode the discrete and flattened $h_{\theta}(u,v,.,.)$. For example, and employing a uniform grid sampling for illustration, $\mM(\theta)_{i,j}=h_{\theta}(i//w,i\%w,j//w,j\%w)$ with $//$ representing the floor division and $\%$ the modulo operation. For the case of multi-channel images, we consider without loss of generality that the same transformation is applied on each channel separately. We depict this operation along with the exact form of $\mM(\theta)$ for the case of the zoom transformation in \cref{fig:summary}.

\subsection{Analytical Expectation and Variance of Transformed Images}
\label{sec:explicit}

The above construction (\cref{def:our_transform}) turns out to make the analytical form of the first two moments of an augmented sample straightforward to derive. As this contributes to one of our core contribution, we propose a step-by-step derivation of $\mathbb{E}_{\rvtheta}\left[ \mathcal{T}_{\rvtheta}(I)\right]$ for the case of horizontal translation (as in \cref{eq:basis_motivation}).

Let's consider again the continuous model (the discrete version is provided after \cref{thm:affine_explicit}). Using Fubini's theorem to switch the order of integration and recalling \cref{eq:basis_image_model}, we have that 
\begin{align}
    \mathbb{E}_{\rvtheta}\left[ \mathcal{T}_{\rvtheta}(I)(u,v)\right]=\hspace{-0.1cm}\int\hspace{-0.05cm} I(x,y) \mathbb{E}_{\rvtheta}\left[h_{\rvtheta}(u,v,x,y)\right]\hspace{-0.05cm}dx dy.\label{eq:expected}
\end{align}
Using the definition for $h$ from \cref{cor:affine} for horizontal translation, $\mathbb{E}_{\rvtheta}\left[h_{\theta}(u,v,x,y)\right]$ becomes
\begin{align*}
    \mathbb{E}_{\rvtheta}\left[\delta(u\hspace{-0.04cm}-\hspace{-0.04cm}x\hspace{-0.04cm}-\hspace{-0.04cm}\theta,v\hspace{-0.04cm}-\hspace{-0.04cm}y)\right]=&\hspace{-0.04cm}\int\hspace{-0.04cm} \delta(u-x-\theta,v-y) p (\theta) d\theta\\
    =& p(u-x)\delta(v-y),
\end{align*}
with $p$ the density function of $\rvtheta$ prescribing how the translation parameter is distributed. As a result, in this univariate translation case, the expected augmented image at coordinate $(u,v)$ is given by
\begin{align}
    \mathbb{E}_{\rvtheta}\left[\mathcal{T}_{\rvtheta}(I)\hspace{-0.05cm}(u,v)\right]&=\int I(x,y)p(u-x)\delta(v-y) dxdy\nonumber \\
    &=\int I(x,v)p(u-x) dx,\label{eq:expectation_example}
\end{align}
which can be further simplified into $\mathbb{E}_{\rvtheta}\left[\mathcal{T}_{\rvtheta}(I)(.,v)\right]= I(.,v) \star p$. Hence, the expected translated image is the convolution (on the $x$-axis only) between the original image $I$ and the univariate density function $p$. We formalize this for the transformations of \cref{cor:affine} below.

\begin{theorem}
\label{thm:affine_explicit}
The analytical form of $\mathbb{E}_{\rvtheta}\left[h_{\rvtheta}(u,v,x,y)\right]$, used to obtain the expected transformed image (recall \cref{eq:expected}) is given by
\begin{align}
    p(u-x,v-y)\text{ and }p\left(\frac{u}{x}\right)\delta\Big(\frac{u}{x}-\frac{v}{y}\Big),\label{eq:expected_h}
\end{align}
for translation and rotation, other cases and $\mathbb{E}_{\rvtheta}[\mathcal{T}_{\rvtheta}(\vx_n)\mathcal{T}_{\rvtheta}(\vx_n)^T]$ are deferred to the appendix due to space limitation.
(Proof in Appendix~\ref{proof:affine_explicit}.)
\end{theorem}
\vspace{-0.1cm}

Notice for example how  the expected image under random $2$-dimensional translations with ($2$-dimensional) density $p$ is given by the convolution $I \star p$, providing a new portal to study and interpret convolutions with nonnegative, sum-to-one filters $p$.
Again, and as per \cref{eq:discrete}, the discretized version of the expected image takes the form of $\mathbb{E}_{\rvtheta}[\vt(\rvtheta)]=\mathbb{E}_{\rvtheta}[\mM(\rvtheta)]\vx$ with the entries of $\mathbb{E}_{\rvtheta}[\mM(\rvtheta)]$ given by discretizing \cref{eq:expected_h}. We depict this expected matrix for various transformations as well as their application onto two different discrete images in \cref{fig:expectation}, and we now proceed on deriving the left-hand side of \cref{eq:original} i.e. the explicit DA regularizer.

\subsection{The Explicit Regularizer of Data-Augmentations}
\label{sec:motivation}

\begin{figure}[t!]
    \centering
    \begin{minipage}{0.025\linewidth}
    \rotatebox{90}{\centering third \hspace{1cm}second \hspace{1cm}first}
    \end{minipage}
    \begin{minipage}{0.23\linewidth}
    \centering
    rotation
    \includegraphics[width=\linewidth]{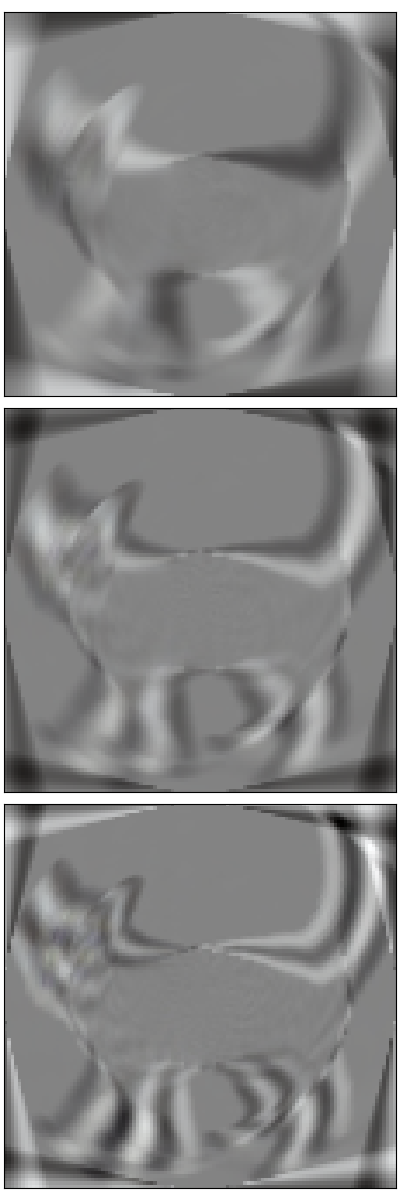}
    \end{minipage}
    \begin{minipage}{0.23\linewidth}
    \centering
    shear
    \includegraphics[width=\linewidth]{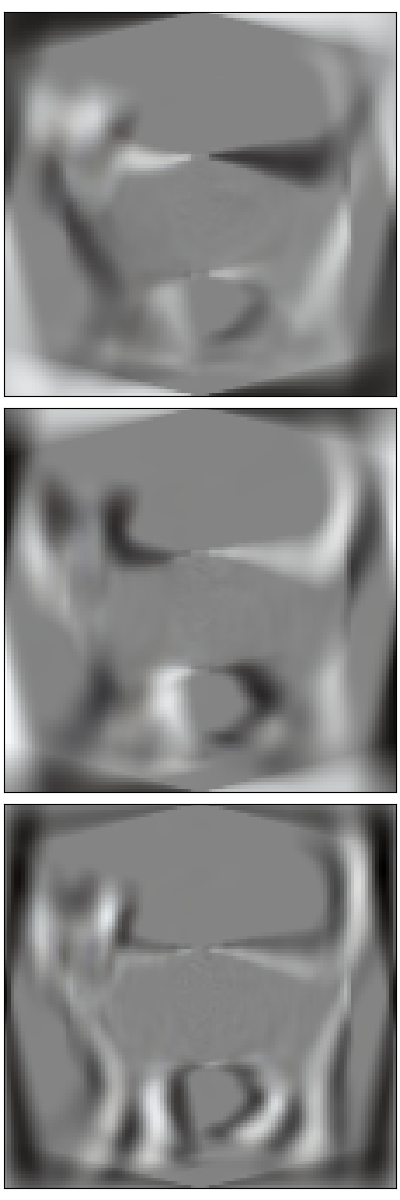}
    \end{minipage}
    \begin{minipage}{0.03\linewidth}
    \rotatebox{90}{rank of $\mathbb{V}[\mathcal{T}(\vx)]$}
    \end{minipage}
    \begin{minipage}{0.44\linewidth}
    \includegraphics[width=\linewidth]{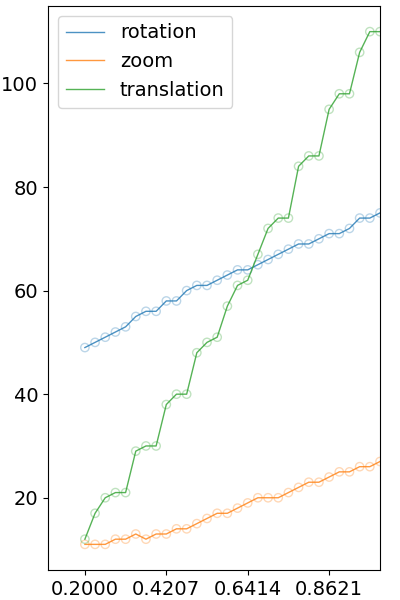}\\
    transformation amplitude
    \end{minipage}
    \vspace{-0.3cm}
    \caption{{\bf Left:}Eigenvectors ($\mQ$ matrix in \cref{thm:explicit}) associated to the largest eigenvalues of the sample variance matrix $\mathbb{V}[\mathcal{T}(\vx)]$ for the rotation and shear augmentations with $\mathcal{U}(-15^\circ,15^\circ)$ and $\mathcal{U}(-15^\circ,15^\circ)\otimes\mathcal{U}(-15^\circ,15^\circ)$ respectively. As per \cref{thm:delta_method}, during training, aligning the kernel of the model's Jacobian matrix (evaluated at each expected sample) to the largest eigenvectors of $\mathbb{V}[\mathcal{T}(\vx)]$ (as seen above) will reduce the loss variance under the respective DA sampling. {\bf Right:} the number of nonzero eigenvalues of $\mathbb{V}[\mathcal{T}(\vx)]$ (nonzero elements in $\Lambda$ in \cref{thm:explicit}) for increasing transformation amplitude (the datum dimension is $114\times 114 \times 3$). We observe that even for high amplitude transformations, the constraint of \cref{eq:delta_ineq} focus on a small umber of dimensions as standard DA span a low-dimensional subspace around each image, increasing linearly with the amplitude of the transformation.}
    \label{fig:eigenvectors}
\end{figure}

\begin{figure}[t!]
    \centering
    \includegraphics[width=0.49\linewidth]{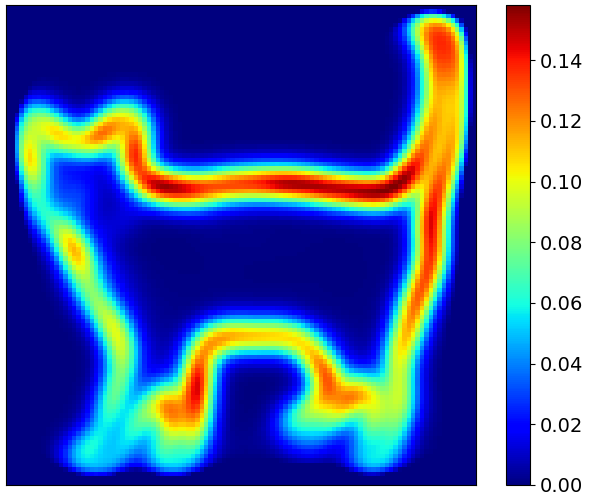}
    \includegraphics[width=0.49\linewidth]{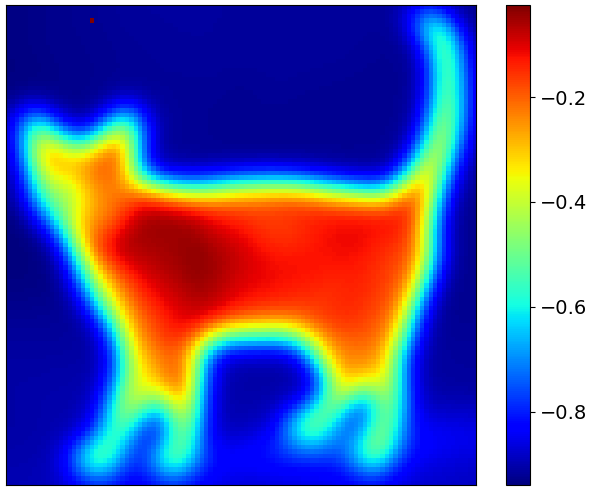}
    \vspace{-0.3cm}
    \caption{Pixel variance i.e. diagonal of $\mathbb{V}[\mathcal{T}(\vx)]$ ({\bf left}) and pixel covariance between a background pixel and all other pixels, i.e. $700^{\rm th}$ row of $\mathbb{V}[\mathcal{T}(\vx)]$ ({\bf right}) reshaped as an image for the cat of \cref{fig:expectation}, for $\mathcal{T}$ translations with distribution $\mathcal{N}(0,0.1)\otimes\mathcal{N}(0,0.1)$. As per \cref{thm:explicit}, the variance of the pixel values, seen on the left, being much higher for the edges of the cat than for its interior body/texture means that the parameters, during training, will be focusing on the cat texture, a phenomenon empirically observed in deep networks \cite{geirhos2018imagenettrained}.}
    \label{fig:variance}
\end{figure}

To keep notations as light as possible, we (for now) consider a linear regression model  with Mean Squared Error (MSE). In that setting, the expected loss under DA sampling (recall \cref{eq:original}) becomes
\begin{equation}
    \mathcal{L} = \sum_{n=1}^{N}\mathbb{E}_{\rvtheta}\left[\left\| \vy_n-\mW \mathcal{T}_{\rvtheta}(\vx_n)-\vb \right\|_2^2 \right],\label{eq:mse_loss}
\end{equation}
with $\vx_n \in \mathbb{R}^{D},n=1,\dots,N$ the input (flattened) n$^{\rm th}$ image $I_n$, $\vy_n \in \mathbb{R}^{K}$ the n$^{\rm th}$ target vector, and $\mW \in \mathbb{R}^{K \times D},\vb \in \mathbb{R}^{K}$ the model's parameters. Recall from \cref{sec:monte_carlo_motivation} that to learn $\mW,\vb$ under DA, the current method consists in performing a one-sample MC estimation. Instead, let's derive (detailed derivation in \cref{sec:derivation}) the exact loss of \cref{eq:mse_loss} as a function of the sample mean and variance under the consider DA. We will drop the $\rvtheta$ subscript for clarity to obtain $\mathcal{L}$ as
\begin{equation}
     \sum_{n=1}^{N}\hspace{-0.05cm}\|\vy_n\hspace{-0.05cm}-\hspace{-0.05cm}\mW\mathbb{E}\left[ \mathcal{T}(\vx_n)\right]\hspace{-0.05cm}-\hspace{-0.05cm}\vb\|_2^2
    \hspace{-0.05cm}+\hspace{-0.05cm}\|\mW\mQ(\vx)\Lambda(\vx)^{\frac{1}{2}}\|_F^2,\label{eq:explicit}
\end{equation}
with the spectral decomposition  $\mQ(\vx)\Lambda(\vx)\mQ(\vx)^T=\mathbb{V}\left[\mathcal{T}(\vx)\right]$. The right term in \cref{eq:explicit} is the explicit DA regularizer. It pushes the kernel space of $\mW$ to align with the largest principal directions of the data manifold tangent space, as modeled by the DA. In fact, the largest eigenvectors in $\mQ(\vx)$ represent the principal directions of the data manifold tangent space at $\vx$, as encoded via $\mathbb{V}\left[\mathcal{T}(\vx)\right]$.

We propose in \cref{fig:eigenvectors} visualization of $\mQ$ and $\Lambda$ for different DAs, illustrating how each DA policy impacts the model's parameter $\mW$ through the regularization of \cref{eq:explicit}. The knowledge of $\mathbb{E}\left[\mathcal{T}(\vx)\right]$ and $\mathbb{V}\left[\mathcal{T}(\vx)\right]$ from \cref{thm:affine_explicit} finally enables to train a (linear) model on the true expected loss (\cref{eq:explicit}) as we formalize below.

\begin{theorem}
\label{thm:explicit}
Training a linear model with MSE and infinite DA sampling is equivalent to minimizing \cref{eq:explicit} and produces the optimal $\mW^*$ model's parameter
\begin{align*}
    \big(\sum_n (\vy_n-\vb)\mathbb{E}[\vx_n]^T\hspace{-0.04cm}\big)\big(\hspace{-0.04cm}\sum_n \mathbb{E}\left[\vx_n\right]\mathbb{E}\left[\vx_n\right]^T\hspace{-0.08cm}+\hspace{-0.05cm}\mathbb{V}[\vx_n]\hspace{-0.05cm}\big)^{-1}\hspace{-0.05cm},
\end{align*}
with $\mathbb{E}(\vx)\triangleq \mathbb{E}[\mathcal{T}(\vx_n)]$ and $\mathbb{V}(\vx)\triangleq \mathbb{V}[\mathcal{T}(\vx_n)]$. We visualize $\mathbb{V}[\vx_n]$ for the translation DA in \cref{fig:variance}.
\end{theorem}
\vspace{-0.2cm}

The same line of result can be derived in the nonlinear setting by assuming that the DA is restricted to small transformations. In that case, one leverages a truncated Taylor approximation of the nonlinear model\footnote{as commonly done, see e.g. Sec.~A.2 from \citet{wei2020implicit} for justification and approximation error analysis} and recovers that (local) DA applies the same regularization than in \cref{eq:explicit} but with the model's Jacobian matrix $\mJ f_{\gamma}(\vx_n)$ in-place of $\mW$ (more details in \cref{sec:variance}).

As we are now in possession of the exact expected loss, we are able to measure precisely how accurate were the MC estimate commonly used to train models under DA sampling which we propose to do in the following section.

\section{Data-Augmentation Sampling Efficiency and Loss Sensitivity}
\label{sec:convergence}

In this section we present the empirical convergence of the loss MC estimate against the true expected loss (\cref{sec:monte_carlo}), and we provide exact variance analysis of that estimate as a function of the model's Jacobian matrix and the sample variance eigenvectors (\cref{sec:sensitivity}).

\subsection{Empirical Monte-Carlo Convergence of Transformed Images} 
\label{sec:monte_carlo}

Now equipped with the closed-form formula for the {\em average image} and {\em average loss} under different image transformation distributions, we propose to empirically measure how efficient is the MC estimation to estimate the exact loss at-hand (recall \cref{eq:original}).

\begin{figure}[t!]
    \centering
    \begin{minipage}{0.24\linewidth}
    \centering
    {\scriptsize original image}
    \end{minipage}
    \begin{minipage}{0.24\linewidth}
    \centering
    {\scriptsize expected\\[-0.5em]$E_{\rvtheta}[T_{\rvtheta}(\vx)]$}
    \end{minipage}
    \begin{minipage}{0.24\linewidth}
    \centering
    {\scriptsize estimated\\[-0.5em]$(N=100)$}
    \end{minipage}
    \begin{minipage}{0.24\linewidth}
    \centering
    {\scriptsize estimated\\[-0.5em]$(N=10000)$}
    \end{minipage}
    \centering
    \includegraphics[width=\linewidth]{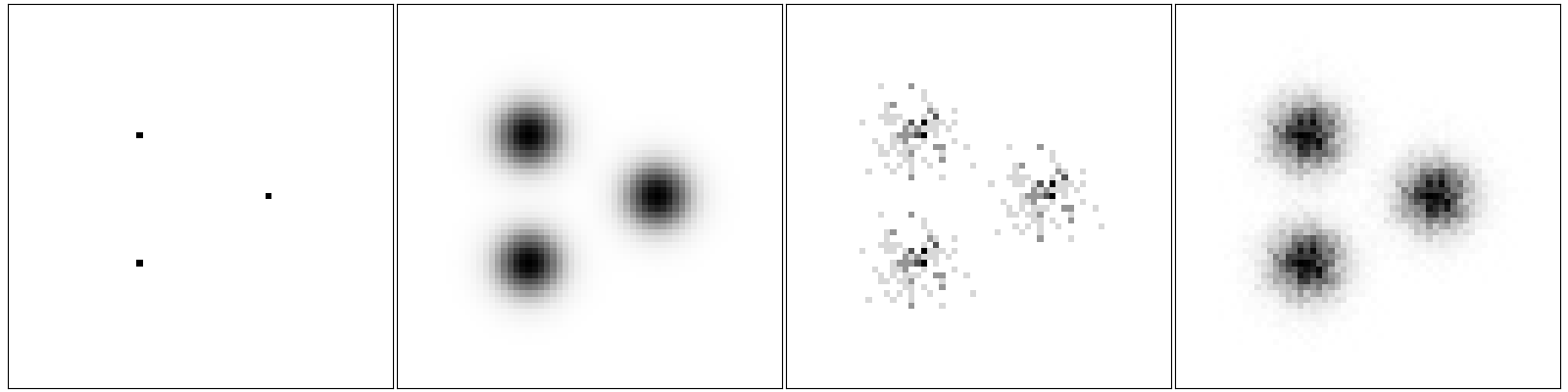}\\
    {\tiny $\ell_2$ between MC estimate and the expected image (left) and the expected MSE loss (right)}
    \begin{minipage}{0.49\linewidth}
    \centering
    \includegraphics[width=\linewidth]{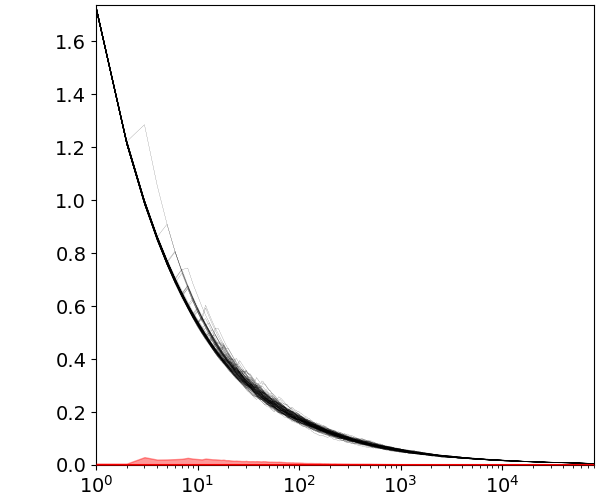}\\[-0.5em]
    {\small $N$ (log-scale)}
    \end{minipage}
    \begin{minipage}{0.49\linewidth}
    \centering
    \includegraphics[width=\linewidth]{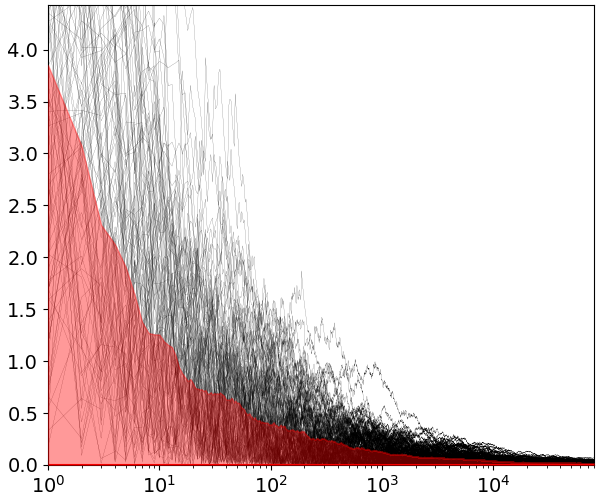}\\[-0.5em]
    {\small $N$ (log-scale)}
    \end{minipage}
        \vspace{-0.3cm}
    \caption{Depiction of the analytical expected image under translation ($\mathcal{N}(0,0.1)\otimes \mathcal{N}(0,0.1)$) against its $N$-sample Monte-Carlo estimate ({\bf top row}), $\ell_2$ distance between the true and estimated images ({\bf bottom left}) and the true and estimated MSE loss with a random Gaussian $\vy,\mW$ ({\bf bottom right}). In red is depicted the standard deviation of the independent Monte-Carlo runs. Clearly we observe that even on a simple $(64\times 64)$ image and using the translation transformation, thousands of sampled images are necessary to provide an accurate estimate of the loss at hand. That is, sampling based data-augmentations are a rather inefficient medium to employ for injecting prior information into a model as this prior information will only emerge after tens of thousands of images have been sampled.}
    \label{fig:convergence_translation}
\end{figure}

\begin{figure}[h!]
    \centering
        \includegraphics[width=\linewidth]{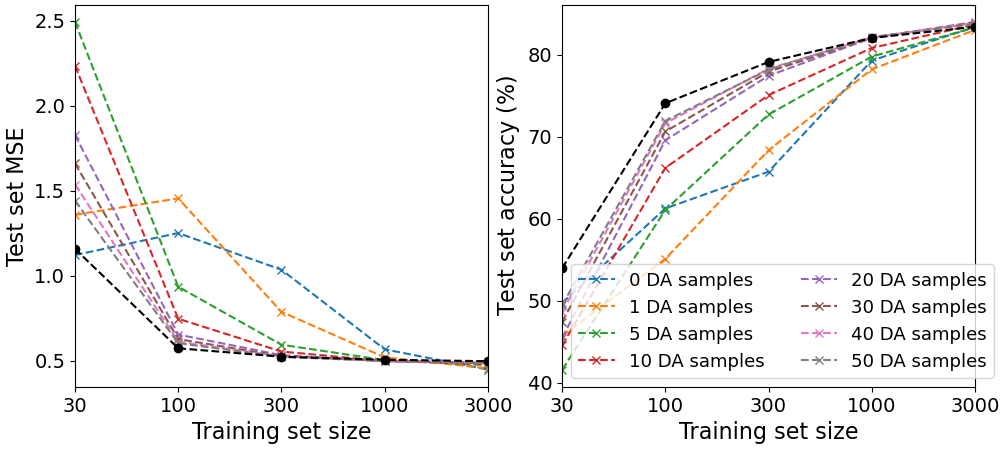}
        \vspace{-0.85cm}
    \caption{Linear regression task on the MNIST dataset with controlled training set size and translation DA sampling against the use of the expected loss ({\bf black line}) for varying number of DA samples ({\bf colored lines}). Evaluation is done on the one-hot encoded labels with the MSE ({\bf top}) and on the accuracy performance ({\bf bottom}) on the full MNIST test set. We observe that in low training data regime even a $50 \times$ increase in dataset size obtained for DA sampling is not enough to provide the performances reached by employing the analytical expected loss. As the training size increases, as the DA samples get redundant with the added samples closing the performance gap.}
    \label{fig:convergence_translation2}
\end{figure}

We first propose in \cref{fig:convergence_translation} a constructed $(64 \times 64)$ image for which we compute the expected loss (\cref{eq:explicit}) and the Monte-Carlo estimate (right-hand side of \cref{eq:original}). Surprisingly, we obtain that even for a simple augmentation policy such as local translations, between 1000 and 10000 samples are required to correctly estimate the MSE loss from the augmented samples. In a more practical scenario, one could rightfully argue that the combination of the DA samples from different images allows to obtain a better estimate with a smaller amount of augmentations per sample. Hence, we provide in \cref{fig:convergence_translation2} that experiment using a linear model on MNIST with varying train set size. We observe that as the number of samples grows as the required number of augmentation per sample reduces. Nevertheless, even with thousands of samples, at least $50$ augmentations per sample are required to provide an accurate estimate.

We now propose to specifically quantify the sensitivity of the MC estimate to DA sampling.

\vspace{-0.2cm}

\subsection{Loss Sensitivity Under Data-Augmentation Sampling in the Linear and Nonlinear Regime}
\label{sec:sensitivity}

Recalling \cref{sec:monte_carlo_motivation}, current DA training is performed on a MC estimate of the loss. The estimator's variance \citep{newman1999monte} is proportional to the variance of the quantity being estimated: $\mathbb{V}[(\mathcal{L}\circ f)(\mathcal{T}(\vx))]$.
In this section, our goal is to characterize that variance precisely to understand when, and why, would an MC estimator converge for a given model.

By leveraging the delta method \citep{doob1935limiting,oehlert1992note} i.e. a truncated Taylor expansion of the model and loss function mapping as $\mathcal{L} \circ f$, we have
\begin{align}
    \mathbb{V}[(\mathcal{L}\circ f)(\mathcal{T}(\vx))]\approx& \|\nabla(\mathcal{L}\circ f)(\mathbb{E}[\mathcal{T}(\vx)])\|^2_{ \mathbb{V}[\mathcal{T}(\vx)]},\label{eq:delta_method}
\end{align}
with $\|\vu\|^2_{\mA}\triangleq \vu^T\mA\vu$. Noticing that $\mathbb{V}[\mathcal{T}(\vx)]=\mathbb{E}[\mathcal{T}(\vx) \mathcal{T}(\vx)^T]-\mathbb{E}[\mathcal{T}(\vx)]\mathbb{E}[\mathcal{T}(\vx)]^T$ and using the closed-form moments of augmented samples from \cref{thm:affine_explicit}, it is possible to write out \cref{eq:delta_method} explicitly for model and DA specific analysis. We visualize $\mathbb{V}[\mathcal{T}(\vx)]$ in Fig.~\ref{fig:variance}.

{\bf Linear regression case.}~To gain some insight into \cref{eq:delta_method}, let's first consider the linear regression case leading to
\begin{align*}
    \mathbb{V}[(\mathcal{L}\circ f)(\mathcal{T}(\vx))]\approx&\|\vy-\mW\mathbb{E}[\mathcal{T}(\vx)]-\vb\|^2_{\mW \mathbb{V}[\mathcal{T}(\vx)]\mW^T}.
\end{align*}
Hence the estimated loss variance depends on the model predicting the correct output when observing the expected sample $\mathbb{E}[\mathcal{T}(\vx)]$,  and on the smallest right singular vectors of $\mW$ to align with the largest eigenvectors of $\mathbb{V}[\mathcal{T}(\vx)]$, echoing our observation below \cref{eq:explicit}. 

{\bf General case.}~Since the result from \cref{eq:delta_method} holds in the general setting, we can generalize the above observation to the more general setting.

\begin{theorem}
\label{thm:delta_method}
The variance of the loss's MC estimate (recall \cref{eq:original}) for an input $\vx_n$ goes to $0$ if the loss gradient $\nabla \mathcal{L}(f(\mathbb{E}[\mathcal{T}(\vx)]))$ goes to $0$, or if the kernel of the model's Jacobian matrix $\mJ f(\mathbb{E}[\mathcal{T}(\vx)])$ aligns with the largest eigenvectors of the sample variance $\mathbb{V}[\mathcal{T}(\vx)]$.
\end{theorem}

We derive here the proof of the above statement. By starting from \cref{eq:delta_method} and applying the Cauchy-Schwarz inequality two times we obtain
\begin{align*}
    \mathbb{V}[(\mathcal{L}\circ f)(\mathcal{T}(\vx))] &\leq \|\nabla \mathcal{L} (f(\mathbb{E}[\mathcal{T}(\vx)])) \|_2^4\\
    \times&\|\mJ f(\mathbb{E}[\mathcal{T}(\vx)])\mathbb{V}[\vx]\mJ f(\mathbb{E}[\mathcal{T}(\vx)])^T\|_F^2,
\end{align*}
and then, rewriting $\mJ f(\mathbb{E}[\mathcal{T}(\vx)]) \mathbb{V}[\mathcal{T}(\vx)\mJ f(\mathbb{E}[\mathcal{T}(\vx)])^T$ as $BB^T$ and using $\|BB^T\|_F\leq \|B\|_F^2$ we obtain
\begin{align}
    \overbrace{\mathbb{V}[(\mathcal{L}\circ f)(\mathcal{T}(\vx))]}^{\mathclap{\;\;\;\;\;\;\small \text{loss $\circ$ model variance under DA}}} &\leq  \|\overbrace{\nabla \mathcal{L} (f(\mathbb{E}[\mathcal{T}(\vx)])) }^{\mathclap{\small \hspace{2.8cm}\text{loss gradient at model(expected image)}}}\|_2^4 \nonumber\\
    \times& \|\underbrace{\mJ f(\mathbb{E}[\mathcal{T}(\vx)])}_{\mathclap{\hspace{-2.5cm}\small\text{model's jacobian at expected image}}}\underbrace{\mQ(\vx)\Lambda^{\frac{1}{2}}(\vx)}_{\mathclap{\hspace{1.4cm}\small \text{DA data manifold tangent space}}}\|_F^4,\label{eq:delta_ineq}
\end{align}
with $\mathbb{V}[\mathcal{T}(\vx)]=\mQ\Lambda^{\frac{1}{2}}\Lambda^{\frac{1}{2}}\mQ^T$. From \cref{eq:delta_ineq}, and assuming that the model's Jacobian does not collapse to $0$ as the training loss is being minimized, we obtain that as the model's Jacobian matrix kernel space aligns with the eigenvectors associated with the largest eigenvalues of $\mathbb{V}[\mathcal{T}(\vx)]$ as the loss variance diminishes, concluding the proof. From \cref{eq:delta_ineq}, it becomes direct to prove that when employing Resnet architectures \citep{he2016identity}, the loss variance will reduce only from the model minimizing the loss at hand, since the Jacobian matrix is always full-rank, preventing any minimization of $\|\mJ f(\mathbb{E}[\mathcal{T}(\vx)])\mQ(\vx)\Lambda^{\frac{1}{2}}(\vx)\|_F^4$.
\\
{\bf Limitations.}~We shall highlight that the results of this section rely on the delta method to approximate the intractable left-hand side of \cref{eq:delta_method} by replacing $(\mathcal{L} \circ f)$ with its Taylor expansion at each $\mathbb{E}[\mathcal{T}(\vx)]$, truncated to the first two terms. However, this has been a good-enough approximation technique in many recent results that have been validated both theoretically and empirically (see footnote on page  \cpageref{fig:variance}).

\vspace{-0.2cm}
\subsection{Explicit Loss Sensitivity Minimization Provably Recovers TangentProp}
\label{sec:variance}

The expected loss (\cref{eq:original}) has been derived in the linear regression setting (\cref{eq:explicit}). But in a more general scenario, and as discussed below \cref{thm:explicit}, this expectation might not be tractable and thus needs to be approximated e.g. based on a Taylor expansion of the loss and model, might be needed. Alternatively to using the approximated expectation, one could employ the usual MC estimate of the expectation (right-hand side of \cref{eq:original}), and leverage the MC estimator variance obtained in \cref{eq:delta_ineq} as a regularizer. We show here that both approaches are equivalent, and recovers a popular regularizer known as TangentProp \citep{simard1991tangent}. Although various extensions of TangentProp have been introduced \citep{chapelle2001vicinal,rifai2011manifold} no principled derivation of it have been yet proposed under the viewpoint of expectation approximation, as we propose below.
Using the same argument as in \cref{sec:sensitivity}, the expectation of a nonlinear transformation of a random variable ($\mathcal{T}(\vx_n)$) can be approximated from the expectation of the Taylor expansion (detailed derivations in \cref{sec:taylor})
\begin{align*}
    \mathbb{E}[(\mathcal{L}&\circ f)(\mathcal{T}(\vx))]\approx(\mathcal{L}\circ f)(\mathbb{E}[\mathcal{T}(\vx)])\\
    &\;\;+\frac{1}{2}\|\underbrace{\mU^{\frac{1}{2}}(\vx)\mV(\vx)^T}_{\mathclap{\text{loss curvature at the model's output}}}\mJ f(\mathbb{E}[\mathcal{T}(\vx)])\mQ(\vx)\Lambda(\vx)^{\frac{1}{2}}\|_F^2,
\end{align*}
with the spectral decomposition of the Hessian matrix  $\mV(\vx)\mU(\vx)\mV^T(\vx)=H \mathcal{L}(f(\mathbb{E}[\mathcal{T}(\vx)])$. Leveraging the Cauchy-Schwarz inequality on the Frobenius norm we finally obtain the following loss upper-bound
\begin{align*}
     (\mathcal{L}\circ f)(\mathbb{E}[\mathcal{T}(\vx)])+\kappa(\vx)\underbrace{\|\mJ f(\mathbb{E}[\mathcal{T}(\vx)])\mQ(\vx)\Lambda(\vx)^{\frac{1}{2}}\|_F^2}_{\mathclap{\text{TangentProp regularization}}},
\end{align*}
with $\kappa(\vx)\geq 0$. As a result, the TangentProp regularizer naturally appears when using a Taylor approximation of the expected loss, and it corresponds to adding an explicit loss variance regularization term (compare the TangentProp with \cref{eq:delta_ineq}).
We thus obtained from first principles that TangentProp emerges naturally when considering the second order Taylor approximation of the expected loss given a DA.

%Another important observation from the TangentProp regularizer, or \cref{eq:delta_ineq} in general, is that it provides a principled interpretation of deep networks saliency maps i.e. the rows of the model's Jacobian matrix. In fact, visualizing the saliency maps of a model provides one way to visualize the important of different parts of an image into the forming of the prediction. However, based on our analysis, we obtain that those saliency maps should be compared against image transformations, as a way to measure the loss sensitivity to that specific transformation. As per \cref{eq:delta_ineq} the squared amplitude of the inner product between the model's saliency maps, and a displacement vector representing an image transformation to be tested, directly measures the loss sensitivity of the current input, to that specified transformation.

\vspace{-0.3cm}

\section{Conclusions}
\vspace{-0.2cm}
In this paper, we proposed a novel set of mathematical tools (\cref{sec:new_transformation,sec:explicit}) under which it is possible to study DA and to provably answer some of the remaining open questions around the efficiency and impact of DA to train a model. We first obtained the explicit regularizer produced by different DAs in \cref{sec:motivation}. This led to the following observation: the kernel space of the $\mW$ matrix is pushed to align with the largest eigenvectors of the sample covariance matrix. This was then studied in a more general setting in \cref{sec:sensitivity} for nonlinear models when characterizing the loss variance under DA sampling. We also observed that Monte-Carlo sampling of transformed images is highly inefficient (\cref{sec:monte_carlo}), even if similar training samples combine their underlying information within a dataset. Lastly, those derivations led us to provably derive a known regularizer -TangentProp- as being the natural minimizer of a model's loss variance. We hope that the proposed analysis will inspire many future works e.g. provable deep network architecture design.

\bibliography{biblio}
\bibliographystyle{icml2022}

\appendix
\onecolumn

\begin{center}
\Huge{Appendix:\\A Data-Augmentation Is Worth A Thousand Samples}
\end{center}

The appendix provides additional supporting materials such as figures, proofs, and additional discussions.

\section{Additional Figures}

\begin{figure}[h!]
    \centering
    \includegraphics[width=0.4\linewidth]{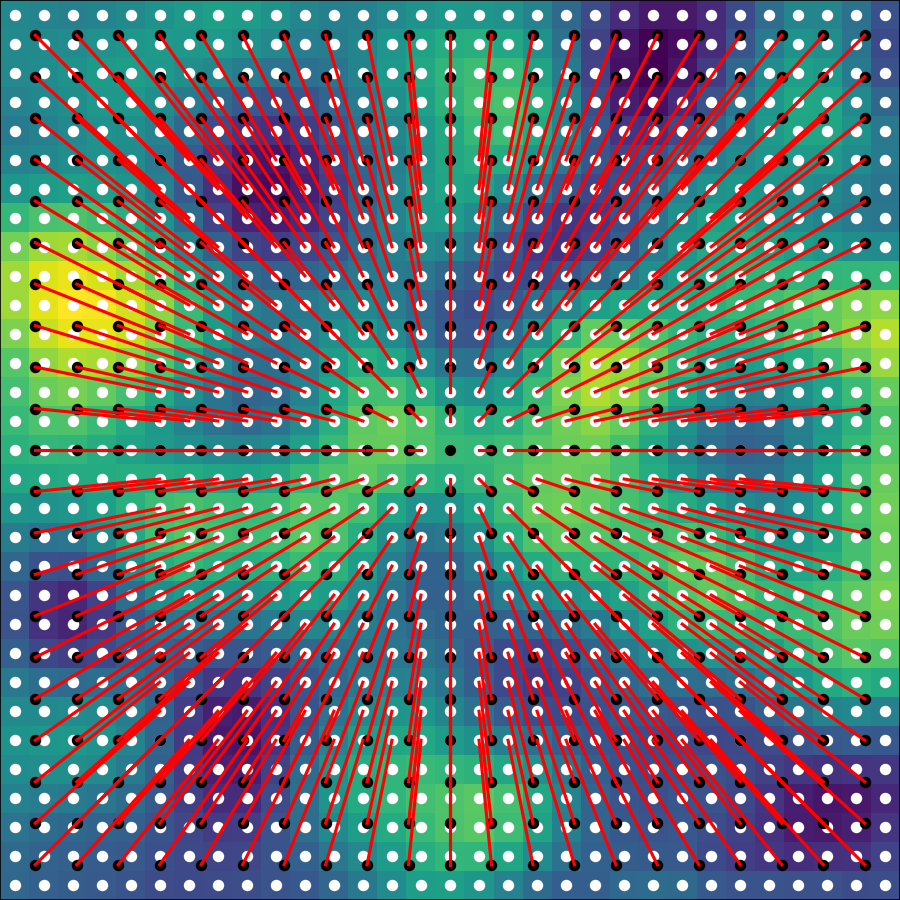}
    \caption{Depiction of the coordinate space transformation applied on a random image (white noise convolved with a low-pass filter. In this setting, one starts from the original image coordinates (white dots) and uses the mapping $t$ (recall \cref{eq:coordinate_transform}) to know what coordinate $t(u,v)$ is associated to the target image coordinate $(u,v)$. In this example of a zoom transformation, the $t$ mapping will scale down those coordinates leading to a displacement (red lines) producing the black dots from the white ones. Once the new coordinates ($t(u,v)$) are known, the actual value is obtained through some flavors of interpolation between the white dots that are the closest to each black dot. For example, using the pixel value of the closest white dot corresponds to a nearest neighbor interpolation. Commonly, the 4 closest pixels are used with a bilinear weighting.}
    \label{fig:coordinate}
\end{figure}

We first propose a visual depiction of coordinate space image transformations in \cref{fig:coordinate}. We employ here the case of a zoom transformation on a randomly generated image for illustration purposes. It can be seen that the transformation acts by altering the position of the image coordinates, and then produce the estimated value of the image at those new coordinates through some interpolation schemes. The most simple strategy is to employ a nearest-neighbor estimate i.e. given the new coordinate, predict the value of the new image at that position to be the same as the pixel value of the closest coordinate. As the coordinate transformation is composed with the image interpolation scheme, it is intricate to obtain any analytical quantity such as the expected image, motivating our proposed pixel-space transformation.

\section{Additional Data-Augmentations}
\label{sec:cutmix}

The case of CutOut, CutMix or MixUp offer interesting avenues to our work. First, we shall highlight that those transformation perform both an alteration of the input $\vx$ and of the corresponding output $\vy$. Due to this double effect, they fall slightly out of the scope of this study.

\section{Expected Mean-Squared Error Derivation}
\label{sec:derivation}

In this section, we provide through direct derivations the expression of the expected mean squared error, under a given data augmentation distribution, as a function of the image first two moments. This result is crucial as it demonstrates that knowledge of those two moments is sufficient to obtain a closed form solution of the expected loss. As far as we are aware, this is a new result enabled by the proposed pixel-space transformation.

Using the cylic property of the Trace operator and the fact that $\mathbb{V}[\mathcal{T}(\vx)]=\mathbb{E}[\mathcal{T}(\vx) \mathcal{T}(\vx)^T]-\mathbb{E}[\mathcal{T}(\vx)]\mathbb{E}[\mathcal{T}(\vx)]^T$ we can express the expected MSE loss as the MSE loss between the target and the expected input plus a regularization term as follows
\begin{align*}
    \mathcal{L}(\mW,\vb) =& \sum_{n=1}^{N}\mathbb{E}_{\theta \sim \Theta}\left[\left\| \vy_n-\mW \mathcal{T}_{\theta}(\vx_n)-\vb \right\|_2^2 \right]\\
    =&\sum_{n=1}^{N}\|\vy_n-\vb\|_2^2-2\langle \mW^T(\vy_n-\vb),\mathbb{E}_{\theta \sim \Theta}\left[ \mathcal{T}_{\theta}(\vx_n)\right]\rangle+\mathbb{E}_{\theta \sim \Theta}\left[\mathcal{T}_{\theta}(\vx_n)^T\mW^T\mW\mathcal{T}_{\theta}(\vx_n)\right]\\
    =&\sum_{n=1}^{N}\|\vy_n-\vb\|_2^2-2\langle \mW^T(\vy_n-\vb),\mathbb{E}_{\theta \sim \Theta}\left[ \mathcal{T}_{\theta}(\vx_n)\right]\rangle+\mathbb{E}_{\theta \sim \Theta}\left[{\rm Tr}\left(\mW^T\mW\mathcal{T}_{\theta}(\vx_n)\mathcal{T}_{\theta}(\vx_n)^T\right)\right]\;\;\text{(cyclic prop.)}\\
    =&\sum_{n=1}^{N}\|\vy_n-\vb\|_2^2-2\langle \mW^T(\vy_n-\vb),\mathbb{E}_{\theta \sim \Theta}\left[ \mathcal{T}_{\theta}(\vx_n)\right]\rangle+{\rm Tr}\left(\mW^T\mW\mathbb{E}_{\theta \sim \Theta}\left[\mathcal{T}_{\theta}(\vx_n)\mathcal{T}_{\theta}(\vx_n)^T\right]\right)\;\;\text{(linear prop.)}\\
    =&\sum_{n=1}^{N}\|\vy_n-\vb\|_2^2+\langle \mW^T(\vy_n-\vb),\mathbb{E}_{\theta \sim \Theta}\left[ \mathcal{T}_{\theta}(\vx_n)\right]\rangle\\
    &+{\rm Tr}\left(\mW^T\mW\left(\mathbb{E}_{\theta \sim \Theta}\left[\mathcal{T}_{\theta}(\vx_n)\mathcal{T}_{\theta}(\vx_n)^T\right]-\mathbb{E}_{\theta \sim \Theta}\left[\mathcal{T}_{\theta}(\vx_n)\right]\mathbb{E}_{\theta \sim \Theta}\left[\mathcal{T}_{\theta}(\vx_n)\right]^T\right)\right)\\
    &+{\rm Tr}\left(\mW^T\mW\left(\mathbb{E}_{\theta \sim \Theta}\left[\mathcal{T}_{\theta}(\vx_n)\right]\mathbb{E}_{\theta \sim \Theta}\left[\mathcal{T}_{\theta}(\vx_n)\right]^T\right)\right)\\
    =&\sum_{n=1}^{N}\|\vy_n-\vb\|_2^2+\langle \mW^T(\vy_n-\vb),\mathbb{E}_{\theta \sim \Theta}\left[ \mathcal{T}_{\theta}(\vx_n)\right]\rangle+{\rm Tr}\left(\mW^T\mW\mathbb{E}_{\theta \sim \Theta}\left[\mathcal{T}_{\theta}(\vx_n)\right]\mathbb{E}_{\theta \sim \Theta}\left[\mathcal{T}_{\theta}(\vx_n)\right]^T\right)\\
    &+{\rm Tr}\left(\mW^T\mW\mathbb{V}_{\theta \sim \Theta}\left[\mathcal{T}_{\theta}(\vx_n)\right]\right)\\
    =&\sum_{n=1}^{N}\|\vy_n-\mW\mathbb{E}_{\theta \sim \Theta}\left[ \mathcal{T}_{\theta}(\vx_n)\right]-\vb\|_2^2 +{\rm Tr}\left(\mW^T\mW\mathbb{V}_{\theta \sim \Theta}\left[\mathcal{T}_{\theta}(\vx_n)\right]\right),
\end{align*}
concluding our derivations. Notice how the regularization term acts upon the matrix $\mW$ through the variance of the image under the specified transformation.

\section{Taylor Approximation}
\label{sec:taylor}

In this section we now describe how to exploit a second order Taylor approximation of any loss and/or transformation of the transformed image as a way to obtain an approximated expected loss/output. Without loss of generality we consider here this mapping to be $(\mathcal{L} \circ f)$, i.e. a nonlinear mapping $f$ and a loss function $\mathcal{L}$. The same result applies regardless of those mappings. The approximation will holds mostly for small transformation due to the quadratic approximation of the mapping.

Let's first obtain the second order Taylor expansion of the nonlinear mapping at $\mathbb{E}[\mathcal{T}(\vx)]$ as
\begin{align*}
    (\mathcal{L}\circ f)(\mathcal{T}(\vx))\approx&(\mathcal{L}\circ f)(\mathbb{E}[\mathcal{T}(\vx)])
    +(\mathcal{T}(\vx_n)-\mathbb{E}[\mathcal{T}(\vx)])^T\nabla (\mathcal{L}\circ f)(\mathbb{E}[\mathcal{T}(\vx)])\\
    &+\frac{1}{2}(\mathcal{T}(\vx_n)-\mathbb{E}[\mathcal{T}(\vx)])^TH (\mathcal{L}\circ f)(\mathbb{E}[\mathcal{T}(\vx)])(\mathcal{T}(\vx_n)-\mathbb{E}[\mathcal{T}(\vx)])
\end{align*}
notice that we are required to at least take the second order Taylor expansion as the first order will vanish as soon as we will take the expectation of that approximation. In fact, taking the expectation of the above, we obtain (using the cyclic property of the Trace operator)
\begin{align*}
    \mathbb{E}[(\mathcal{L}\circ f)(\mathcal{T}(\vx))]\approx&(\mathcal{L}\circ f)(\mathbb{E}[\mathcal{T}(\vx)])+\frac{1}{2}{\rm Tr}\left(H (\mathcal{L}\circ f)(\mathbb{E}[\mathcal{T}(\vx)])\mathbb{V}(\mathcal{T}(\vx))\right)\\
    =&(\mathcal{L}\circ f)(\mathbb{E}[\mathcal{T}(\vx)])+\frac{1}{2}{\rm Tr}\left(\mJ f_{\gamma}(\mathbb{E}[\mathcal{T}(\vx)]^TH \mathcal{L}(f_{\gamma}(\mathbb{E}[\mathcal{T}(\vx)])\mJ f_{\gamma}(\mathbb{E}[\mathcal{T}(\vx)]\mathbb{V}(\mathcal{T}(\vx))\right)\\
    =&(\mathcal{L}\circ f)(\mathbb{E}[\mathcal{T}(\vx)])+\frac{1}{2}\|\mU^{\frac{1}{2}}(\vx)\mV(\vx)^T\mJ f_{\gamma}(\mathbb{E}[\mathcal{T}(\vx)])\mQ(\vx)\Lambda(\vx)^{\frac{1}{2}}\|_F^2,
\end{align*}
where $H$ represents the Hessian. The above concludes our derivation that led to a generalization of the linear case. In fact, notice that in the linear with mean squared error the second order Taylor approximation is exact and the above is exactly the same as the one of the previous section.

\section{Proof of Thm.~\ref{thm:affine_explicit}}
\label{proof:affine_explicit}

This section provides the derivation of the first two moments of images under specific image transformations. Note that the actual distribution is abstracted away as simple $p$. In fact, those results do not depend on the specific form of $p$, rather, they depend on the type of transformation being applied e.g. rotation, translation or zoom. We thus propose to derive them, following the same recipe one will be able to obtain the analytical form of the first two moments for any desired transformation. One fact that we will heavily leverage is the fact that integrating a functional and a Dirac function can be expressed as evaluating that function as the position of the Dirac (recall \cref{thm:equi}).

Throughout this proof, we will denote by $T(u,v;\theta)$ the value of the transformed image at spatial position $(u,v)$, hence $\mathbb{E}_{\theta \sim \Theta}[T(u,v;\theta)]$ is the expected value of the transformed image at pixel position $(u,v)$. And $\mathbb{E}_{\theta \sim \Theta}[T(u,v;\theta)T(u',v';\theta)]$ is the second order (uncentered) moment representing the interplay between pixel positions $(u,v)$ and $(u'v')$ of the transformed image. This second order moment and the first order moment can be used to obtained the variance/covariance of the transformed image.
% \url{https://math.stackexchange.com/questions/2876869/two-dirac-delta-functions-in-an-integral}

{\bf Vertical and Horizontal translation:}~The case of vertical and horizontal translations is taken care of jointly, for vertical-only or horizontal-only transformations, simply use a distribution that is a Dirac (at $0$) for the transformation that is not needed. We thus obtain in general given a $2$-dimensional density $p$ as 
\begin{align*}
\mathbb{E}_{\theta \sim \Theta}[T(u,v;\theta)] =& \int_{\theta}p(\theta)\int I(x,y)h_{\theta}(u,v,x,y)dxdyd\theta\\
    =& \int_{-\infty}^{\infty}p(\theta_1,\theta_2)\int I(x,y)\delta(u=x+\theta_1,v=y+\theta_2)dxdyd\theta\\
    =& \int I(x,y)\int_{-\infty}^{\infty}p(\theta_1,\theta_2)\delta(u=x+\theta_1,v=y+\theta_2)d\theta dxdy\\
    =&\int I(x,y)p(u-x,v-y)dxdy
\end{align*}
from this we also directly obtain that the expected image can be expressed as a $2$-dimensional convolution between the original image and the density being employed for the translation as in $\mathbb{E}_{\theta \sim \Theta}[T(.,.;\theta)]=I \star p$. We now derive the second order moments
\begin{align*}
    \mathbb{E}_{\theta \sim \Theta}[T(u,v;\theta)T(u',v';\theta)] =& \int I(x,y)I(x',y')\int_{-\infty}^{\infty}p(\theta_1,\theta_2)\delta(u=x+\theta_1,v=y+\theta_2)\\
    &\times \delta(u'=x'+\theta_1,v'=y'+\theta_2)d\theta_1d\theta_2 dxdydx'dy'\\
    =& \int I(x,y)I(x',y')p(u-x,v-y)\delta(v-y-(v'-y')) \delta(u-x-(u'-x')) dxdx'dydy'\\
    =& \int I(x'-u'+u,y'-v'+v)I(x',y')p(u-x'+u'-u,v-(y'-v'+v))dx'dy'\\
    =& \int I(x'-u'+u,y'-v'+v)I(x',y')p(-x'+u',-y'+v')dx'dy'\\
    =& \int I(a+u,b+v)I(a+u',b+v')p(-a,-b)dadb,
\end{align*}
as a result the second order moment of the image at $(u,v)$ and ($u',v'$) is simply the inner product between the image translated to $(u,v)$, $(u'v')$ and the density $p$.

{\bf Vertical Shear:} we now move on to the vertical shear transformation. Note that this transformation can be seen as a special case of a translate but with a translation coefficient varying with row/columns. We obtain the following derivations
\begin{align*}
\mathbb{E}_{\theta \sim \Theta}[T(u,v;\theta)] =& \int_{\theta}p(\theta)\int I(x,y)h_{\theta}(u,v,x,y)dxdyd\theta\\
    =& \int_{\theta}p(\theta)\int I(x,y)\delta(u=x+\theta*v,v=y)dxdyd\theta\\
    =& \int I(x,y)\int_{-\infty}^{\infty}p(\theta)\delta(u=x+\theta*v,v=y)d\theta\\
    =&\int I(x,v)p(\frac{u-x}{v})dx\\
    \implies \mathbb{E}_{\theta \sim \Theta}[T(.,v;\theta)] =&I(.,v)\star p(\frac{.}{v}),
\end{align*}
as a result we see that an efficient way to obtain the expected image (each row/column of it) is via a $1$-dimensional convolution with the density being rescaled based on the considered row/column. We now consider the second order moment below
\begin{align*}
    \mathbb{E}_{\theta \sim \Theta}[T(u,v;\theta)T(u',v';\theta)] =& \int I(x,y)I(x',y')\int_{-\infty}^{\infty}p(\theta)\\
    &\times \delta(u=x+\theta*v,v=y)\delta(u'=x'+\theta*v',v'=y')dxdydx'dy'd\theta\\
    =& \int I(x,v)I(x',v')\int_{-\infty}^{\infty}p(\theta)\delta(u=x+\theta*v)\delta(u'=x'+\theta*v')dxdx'd\theta\\
    =& \int I(x,v)I(x',v')p(\frac{u-x}{v})\delta(\frac{u-x}{v}=\frac{u'-x'}{v'})dxdx'\\
    =& \int I(x,v)I(x',v')p(\frac{u-x}{v})\delta(x=u+(x'-u')\frac{v}{v'})dxdx'\\
    =& \int I(u+(x'-u')\frac{v}{v'},v)I(x',v')p(\frac{u'-x'}{v'})dx'\\
    =& \int I(u+zv,v)I(u'+zv',v')p(-z)v'dz,
\end{align*}
concluding the shearing transformation results.
For the horizontal shear, simply do the above derivations with the image axes swapped.

{\bf Rotation:}
\begin{align*}
    \mathbb{E}_{\theta \sim \Theta}[T(u,v;\theta)] =& \int_{\theta}p(\theta)\int I(x,y)h_{\theta}(u,v,x,y)dxdyd\theta\\
    =& \int_{\theta}p(\theta)\int I(x,y)\delta(u=\cos(\theta)x-\sin(\theta)y,v=\sin(\theta)x+\cos(\theta)y)dxdyd\theta\\
    =& \int_{\theta}p(\theta)\int I(x,y)\delta(x^2+y^2=u^2+v^2,\theta =\arctan(y/x)-\arctan(v/u))dxdyd\theta\\
    =& \int I(x,y)\int_{\theta}p(\theta)\delta(x^2+y^2=u^2+v^2,\theta =\arctan(y/x)-\arctan(v/u))dxdyd\theta\\
    =&\int I(x,y)p(\arctan(y/x)-\arctan(v/u)) \delta(x^2+y^2=u^2+v^2)dx dy
\end{align*}

\begin{align*}
    \mathbb{E}_{\theta \sim \Theta}[T(u,v;\theta)T(u',v';\theta)]=& \int_{\theta}p(\theta)\int I(x,y)h_{\theta}(u,v,x,y)dxdy\int I(x',y')h_{\theta}(u',v',x',y')dx'dy'd\theta\\
    =& \int I(x,y)I(x',y')\int_{0}^{2\pi}p(\arctan(y/x)-\arctan(v/u))\\
    &\times\delta(\arctan(y/x)-\arctan(v/u)=\arctan(y'/x')-\arctan(v'/u'),\\
    &\;\;x^2+y^2=u^2+v^2,x'^2+y'^2=u'^2+v'^2)dxdydx'dy'
\end{align*}

\iffalse
Oriingal zoom
{\bf Zoom:}
\begin{align*}
    \mathbb{E}_{\theta \sim \Theta}[P(u,v;x,y,\theta)] =& \int_{-\infty}^{\infty}p(\theta)\delta(u=\theta x,v=y\theta)d\theta\\
    =&p(\frac{u}{x})\delta(\frac{u}{x}=\frac{v}{y})
\end{align*}
\begin{align*}
    \mathbb{E}_{\theta \sim \Theta}[P(u,v;x,y,\theta)P(u',v';x',y',\theta)] =& \int_{-\infty}^{\infty}p(\theta)\delta(u=\theta x,v=y\theta)\delta(u'=\theta x',v'=y'\theta)d\theta\\
    =& \int_{-\infty}^{\infty}p(\frac{u}{x})\delta(u/x=v/y=u'/x'=v'/y')d\theta,
\end{align*}

\fi
This concludes our derivations. Note that while we focused here on the most common transformations, the same recipe can be employed to obtain the expected transformed image for more complicated transformations.
\end{document}